\documentclass[sigconf]{acmart}

\usepackage{url} 
\usepackage{hyperref}
\usepackage{xurl}  
\usepackage{graphicx} 
\usepackage{caption}   
\usepackage{booktabs}  
\usepackage{booktabs}
\usepackage{multirow}
\usepackage{makecell}
\usepackage{pifont}
\usepackage{xcolor}
\usepackage{float}     
\usepackage{graphicx}
\usepackage{subcaption}
\usepackage{makecell}
\usepackage{array}
\usepackage{tabularx}

\newcommand{\cmark}{\textcolor{green!60!black}{\ding{51}}}
\newcommand{\xmark}{\textcolor{red!70!black}{\ding{55}}}

\setcopyright{acmlicensed}
\copyrightyear{2025}
\acmYear{2025}
\acmDOI{XXXXXXX.XXXXXXX}

\acmConference[Conference acronym 'XX]{Make sure to enter the correct
  conference title from your rights confirmation email}{June 03--05,
  2018}{Woodstock, NY}

\acmISBN{978-1-4503-XXXX-X/2018/06}

\title{Context-Aware Pseudo-Label Scoring for Zero-Shot \\ Video Summarization}

\author{Yuanli Wu}
\affiliation{%
	\institution{Nanyang Technological University}
	\country{Singapore}}
\email{YUANLI001@e.ntu.edu.sg}

\author{Long Zhang}
\affiliation{%
	\institution{Guilin University of Electronic Technology}
	\city{Guilin}
	\country{China}}
\email{18291041700@163.com}

\author{Yue Du}
\affiliation{%
	\institution{Shenzhen Institute of Advanced Technology, Chinese Academy of Sciences}
	\city{Shenzhen}
	\country{China}}
\email{yue.du2@siat.ac.cn}

\author{Bin Li}
\authornote{Corresponding author}
\affiliation{%
	\institution{Shenzhen Institute of Advanced Technology, Chinese Academy of Sciences}
	\city{Shenzhen}
	\country{China}}
\email{b.li2@siat.ac.cn}

\begin{document}

\begin{abstract}
With the rapid proliferation of video content across social media, surveillance, and education platforms, efficiently summarizing long videos into concise yet semantically faithful surrogates has become increasingly vital. Existing supervised methods achieve strong in-domain accuracy by learning from dense annotations but incur high labeling costs and limited cross-dataset generalization, while unsupervised approaches, though label-free, often miss high-level human semantics and fine-grained narrative cues. Recent zero-shot pipelines leverage large language models (LLMs) for training-free video summarization, yet remain highly sensitive to handcrafted prompts and dataset-specific normalization.To address these limitations, we propose a rubric-guided, pseudo-labeled prompting framework that converts a small subset of human annotations into high-confidence pseudo labels and aggregates them into structured, dataset-adaptive scoring rubrics for interpretable scene evaluation. During inference, first and last segments are scored based solely on their descriptions, whereas intermediate segments incorporate brief contextual summaries of adjacent scenes to assess narrative progression and redundancy. This contextual prompting enables the LLM to balance local salience and global coherence without parameter tuning.Our framework achieves consistent improvements across three benchmarks. On \emph{SumMe} and \emph{TVSum}, it reaches F1 scores of \textbf{57.58} and \textbf{63.05}, surpassing the zero-shot baseline \emph{PROMPTS TO SUMMARIES} (56.73 and 62.21) by \textbf{+0.85} and \textbf{+0.84}, respectively, while approaching the performance of supervised models. On the query-focused \emph{QFVS} benchmark, our method attains an average F1 of \textbf{53.79}, exceeding the baseline (53.42) by \textbf{+0.37} and demonstrating stable performance across all validation videos. These results confirm that rubric-guided pseudo labeling, combined with contextual prompting, effectively stabilizes LLM-based scoring and establishes a general, interpretable zero-shot paradigm applicable to both generic and query-focused video summarization.\textit{Code is available at: \url{https://github.com/wuyuanli60-svg/Context-Aware-Pseudo-Label-Scoring-for-Zero-Shot-Video-Summarization}}.
\end{abstract}

\ccsdesc[500]{Computing methodologies~Video summarization}
\ccsdesc[300]{Computing methodologies~Natural language processing}
\ccsdesc[300]{Computing methodologies~Visual content analysis}

\keywords{Zero-shot Video Summarization, Large Language Models, Prompt Engineering, Pseudo Labeling, Context-Aware Scoring, Interpretability}
\maketitle

\section{Introduction}

With the surge of video content across social media, surveillance, and online education, efficiently understanding and managing long videos under constrained compute and time budgets has become increasingly important \cite{li2023overview, li2024towards, li2024overview, li2025overview}. \textit{Video summarization} mitigates information overload by condensing lengthy videos into concise and semantically faithful surrogates that preserve key narrative elements, thereby enabling fast browsing, retrieval, and storage reduction \cite{gygli2014summe, song2015tvsum}. Public benchmarks such as SumMe \cite{gygli2014summe} and TVSum \cite{song2015tvsum} provide frame- and keyshot-level annotations with standardized protocols for fair comparison and reproducibility. Nevertheless, achieving high semantic fidelity alongside stable behavior and robust cross-dataset generalization remains challenging in practice.
\begin{figure}[t]
  \centering

  \begin{subfigure}{\linewidth}
    \centering
    \includegraphics[width=\linewidth]{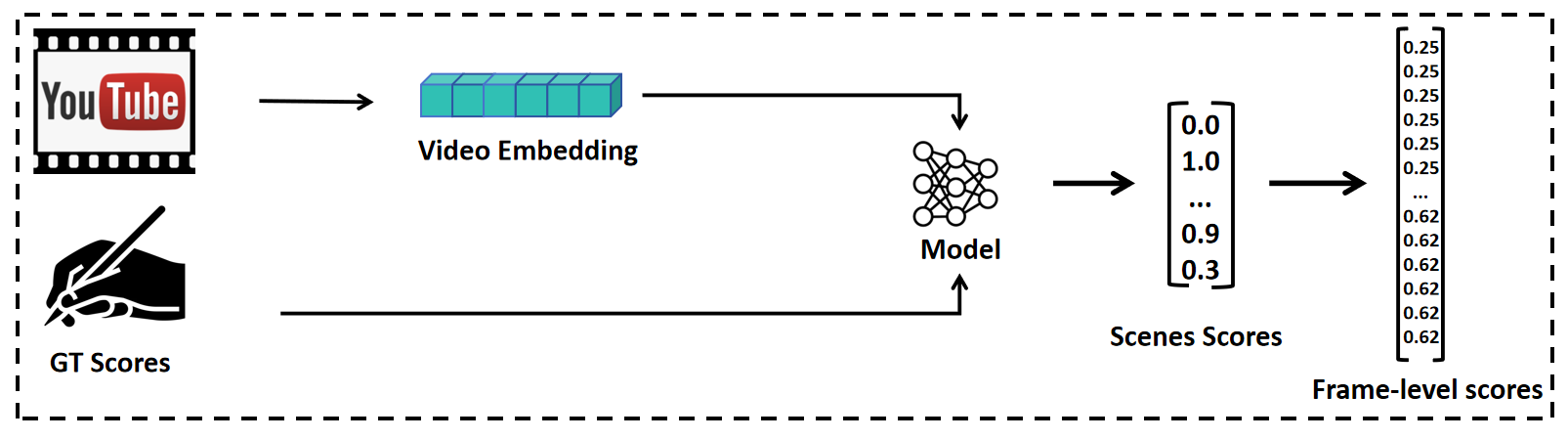}

    \caption{Traditional Supervised/Unsupervised Frame Importance Score Strategy}
  \end{subfigure}


  \begin{subfigure}{\linewidth}
    \centering
    \includegraphics[width=\linewidth]{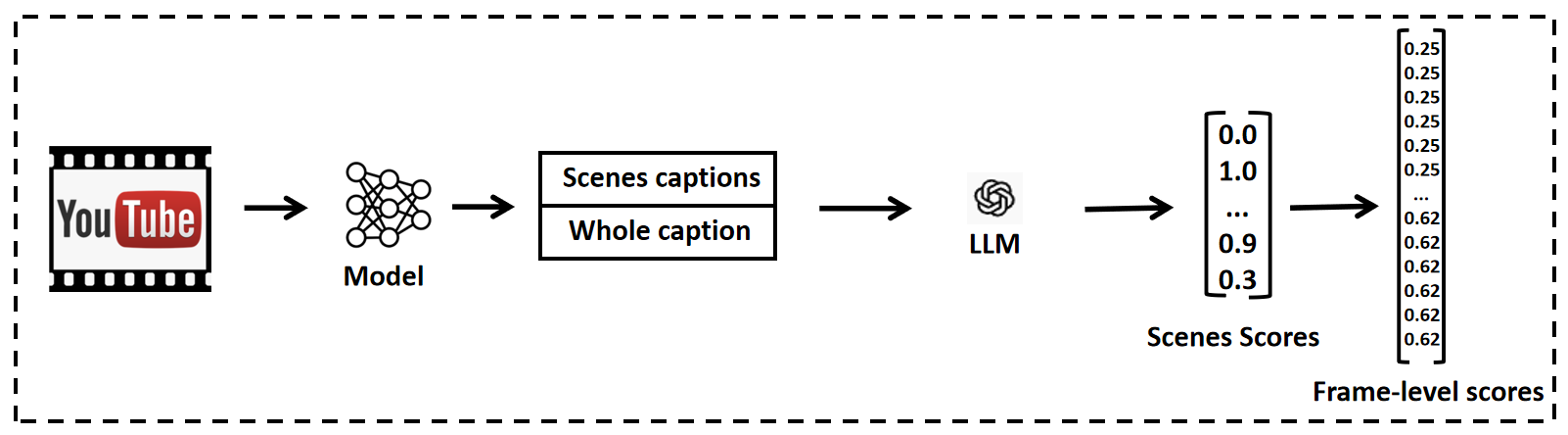}

    \caption{ZERO-SHOT Frame Importance Score Strategy}
  \end{subfigure}


  \begin{subfigure}{\linewidth}
    \centering
    \includegraphics[width=\linewidth]{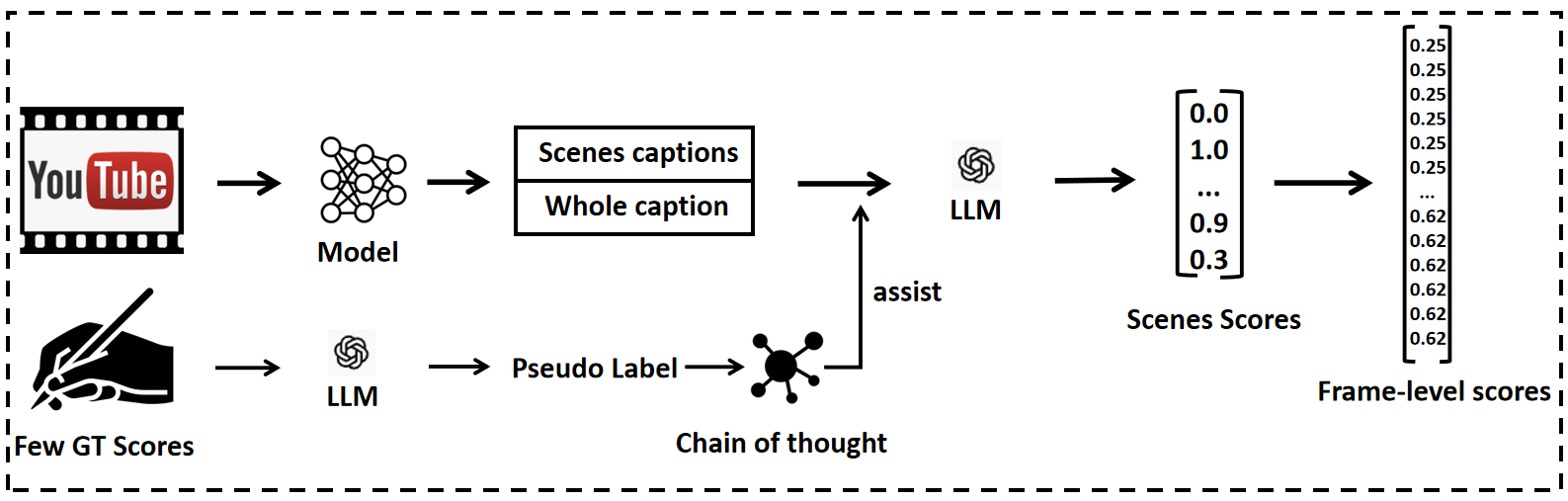}

    \caption{Pseudo Label Optimization ZERO-SHOT Frame Importance Score Strategy}
  \end{subfigure}
  \vspace{-0.6cm}
  \caption{Illustrations of different methods.}
  \label{fig:overview}
  \vspace{-0.5cm}
\end{figure}
Existing studies generally fall into three paradigms. \textbf{Supervised methods} train frame- or shot-level importance predictors using dense annotations, attaining strong in-domain performance but suffering from expensive labeling costs and sensitivity to annotation styles and distribution shifts, which hinder generalization \cite{zhang2016lstm}. \textbf{Unsupervised approaches} optimize objectives such as sparsity, diversity, and representativeness, or employ adversarial/reconstruction surrogates to balance compression and fidelity, avoiding labels yet often failing to capture high-level semantics and user intent required for interpretable summaries \cite{gygli2015submodular, vo2025attention}. More recently, \textbf{prompt-based pipelines} utilize vision–language models to describe scenes and large language models (LLMs) to score scene importance under handcrafted prompts, achieving competitive results without task-specific training \cite{prompts2summaries}. However, these pipelines are highly sensitive to wording and template design, depend on dataset-specific expertise, and require manual normalization (e.g., different score transforms for SumMe vs.\ TVSum), revealing poor dataset adaptivity.

To overcome these issues, we propose a dataset-adaptive prompting paradigm that replaces ad-hoc prompt design with a data-driven pseudo labeling process grounded in a small portion of ground truth. Specifically, representative GT instances and scene descriptions are used to induce \textit{LLM-generated pseudo labels} capturing dataset-specific preferences across narrative coherence, temporal continuity, semantic diversity, and event salience. These pseudo labels are aggregated into a structured rubric that guides automatic prompt synthesis. The rubric-driven prompts are then applied uniformly to all scenes to produce stable scene scores, which are redistributed and smoothed into consistent frame-level importance curves. By structuring the process as ``GT $\rightarrow$ pseudo labels $\rightarrow$ rubric/prompt,'' the approach internalizes dataset scale and style differences into the rubric, reducing reliance on handcrafted templates and post-hoc normalization \cite{prompts2summaries}.

This work makes four main contributions. (1) It introduces a weakly supervised framework that leverages limited GT data to generate LLM-based pseudo labels for guiding rubric and prompt construction, bridging training-free evaluation with dataset adaptivity. (2) It establishes an interpretable and reproducible pipeline from ``GT $\rightarrow$ pseudo labels $\rightarrow$ rubric/prompt $\rightarrow$ scene scores $\rightarrow$ frame mapping/smoothing,'' ensuring stability while balancing narrative coverage, temporal coherence, and diversity. (3) Compared with purely prompting-based zero-shot methods, it markedly reduces prompt sensitivity and dependence on human priors by grounding prompts in learned, dataset-specific criteria \cite{prompts2summaries}. (4) Extensive experiments on \emph{SumMe} \cite{gygli2014summe}, \emph{TVSum} \cite{song2015tvsum}, and the query-focused long-video benchmark \emph{QFVS} \cite{sharghi2017qfvs} demonstrate consistent performance gains and stability. Our framework achieves F1 scores of \textbf{57.58}, \textbf{63.05}, and \textbf{53.79} on SumMe, TVSum, and QFVS, respectively, surpassing the zero-shot baseline by \textbf{+0.85}, \textbf{+0.84}, and \textbf{+0.37}. These results highlight the scalability of rubric-guided pseudo labeling and contextual prompting to both short-form and query-driven long-form video summarization, establishing a unified, interpretable, and training-free paradigm for general video summarization.

\begin{figure*}[t]
  \centering
  \includegraphics[width=\textwidth]{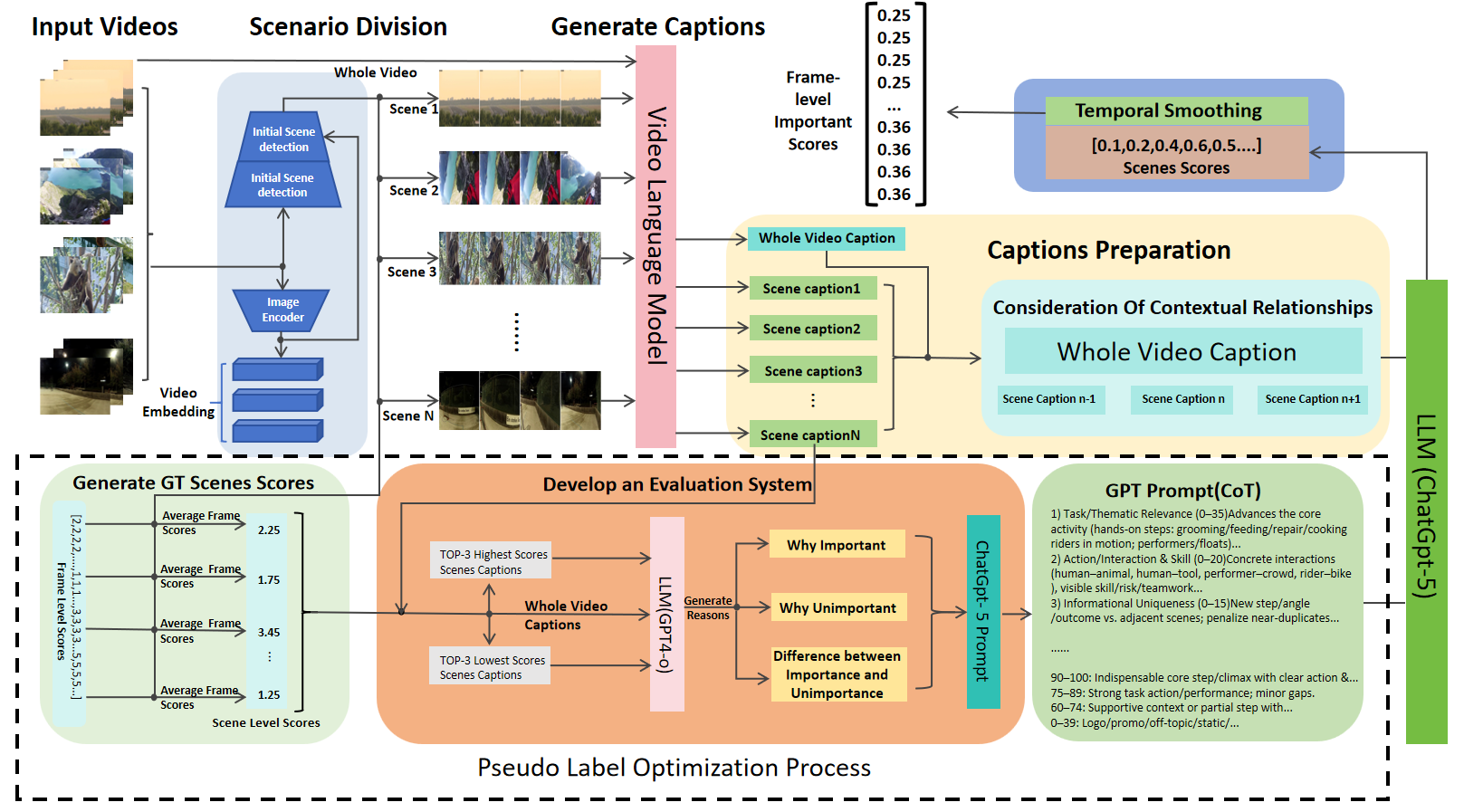}
    \vspace{-0.3cm}
  \caption{Overall architecture of the proposed framework.}
  \label{fig:MethodImplementationProcess}
    \vspace{-0.2cm}
\end{figure*}
  \vspace{-0.3cm}

\section{Related Work}
Video summarization has progressed from supervised sequence models to weaker supervision and prompt-based pipelines with large language models (LLMs). Early work learned frame/shot importance from dense labels and long-range dependencies, showing strong results on SumMe/TVSum \cite{gygli2014summe, song2015tvsum, zhang2016lstm}; later methods reduced annotation with weak/self-supervision and submodular/unsupervised objectives \cite{gygli2015submodular, mahasseni2017sumgan}. 

\textbf{Training-free prompting.}
Recent zero-shot systems describe scenes with vision–language models and rely on LLM prompts to score importance, reaching competitive performance but remaining sensitive to template wording and dataset-specific normalization \cite{prompts2summaries, gygli2014summe, song2015tvsum}.

\textbf{Pseudo labeling.}
Pseudo labels (self-training) improve stability and semantic consistency, historically via confidence- or consistency-based supervision \cite{lee2013pseudo, xie2020noisy, sohn2020fixmatch}, and more recently for prompt/evaluation optimization—e.g., expanding task instructions \cite{wang2023selfinstruct}, rubric-guided judging \cite{zheng2023judge, hashemi2024rubricllm}, and alignment for reliable scoring \cite{zhou2024plabelalign}. In video summarization, pseudo labels encourage relevance and temporal coherence (e.g., \textit{AC-SUM-GAN}, \textit{Cycle-SUM}/\textit{RS-SUM}) \cite{acsumgan, rssum}.

\textbf{Query-focused video summarization (QFVS).}
QFVS conditions summaries on a user query; SH-DPP and the QFVS benchmark established the setting and protocols \cite{sharghi2016qfes, sharghi2017qfvs}, while \textit{CLIP-It} unified generic and query-focused summarization under language guidance \cite{clipit}. Remaining challenges include template sensitivity and dataset adaptivity, motivating our rubric-guided, pseudo-labeled prompting with contextualized scoring.

\textbf{Chain-of-thought (CoT) optimization.}
CoT prompting and its variants provide a practical way to improve LLM reasoning reliability—few/zero-shot CoT \cite{wei2022cot, kojima2022zeroshot} and self-consistency decoding \cite{wang2022selfconsistency} reduce prompt brittleness; for evaluation, LLM-as-a-Judge and rubric-based judging improve consistency and calibration \cite{zheng2023judge, hashemi2024rubricllm}. Our design leverages these insights to stabilize rubric-guided scoring while retaining zero-shot generality.

\section{Method}
Figure~\ref{fig:MethodImplementationProcess} illustrates the overall architecture of our proposed \textbf{rubric-guided zero-shot video summarization framework}, which consists of two major stages: the \textit{pseudo-label optimization stage} and the \textit{inference and scoring stage}. 
In the first stage, the model first divides the input video into a sequence of semantically coherent scenes through the \textbf{Input Videos / Scenario Division} module. Then, a video–language model (\textbf{Video Language Model}) generates natural-language captions for each scene. Based on a small subset of human-annotated examples, scene-level \textbf{pseudo labels} are created (\textbf{Generate GT Scene Scores}). These pseudo labels are further analyzed and abstracted by the large language model (ChatGPT-5) to construct structured, multi-dimensional \textbf{rubrics} and reusable \textbf{prompt templates} (\textbf{Develop Evaluation System / GPT Prompt (CoT)}). 
In the second stage, the model performs zero-shot summarization for unseen videos. For each scene, the \textbf{Captions Preparation} module prepares the textual input: the first and last scenes are evaluated using only their own captions, while intermediate scenes include brief contextual summaries of adjacent scenes and the global description to model narrative continuity. The large language model then scores each scene based on the rubric-guided prompts (\textbf{LLM Scoring}), and a temporal smoothing module (\textbf{Temporal Smoothing}) converts scene-level scores into continuous frame-level importance distributions. 
Overall, this framework establishes a complete pipeline from ``few labeled samples → pseudo labels → rubric/prompt → scene-level scoring → frame-level summarization'', achieving both interpretability and strong cross-dataset generalization.

\subsection{Scenario Division}
We split each video into consecutive, semantically coherent scenes with a lightweight vision-based detector and an adaptive threshold. \textbf{Perceptual hashes are computed using the Python \texttt{ImageHash} library} (\texttt{imagehash.phash}, PIL backend).

\textbf{Step 1. Frame preprocessing.}
Given a frame $I_t$, convert to grayscale and resize to a fixed resolution $N\times N$:
\begin{equation}
X_t \in \mathbb{R}^{N\times N}.
\end{equation}

\textbf{Step 2. Perceptual hash (pHash).}
Apply \texttt{imagehash.phash} to $X_t$ to obtain a binary hash vector $h_t \in \{0,1\}^{M_h}$.

\textbf{Step 3. Frame-wise difference (normalized Hamming).}
The structural change between consecutive frames is
\begin{equation}
D(t,t{+}1) \;=\; \frac{1}{M_h}\sum_{m=1}^{M_h}\mathbb{1}\!\left[h_t^{(m)} \neq h_{t+1}^{(m)}\right].
\label{eq:hamming}
\end{equation}
\textit{Symbols:} $M_h$ is hash length (bits); $h_t^{(m)}$ is the $m$-th bit of $h_t$; $\mathbb{1}[\cdot]$ is the indicator.

\textbf{Step 4. Scene boundary decision.}
Declare a boundary when the change exceeds a threshold:
\begin{equation}
\mathrm{Boundary}(t)=\mathbb{1}\!\left[D(t,t{+}1)\ge \tau\right].
\label{eq:boundary}
\end{equation}

\textbf{Step 5. Adaptive threshold selection.}
Instead of a fixed $\tau$, select $\tau^\star$ on a grid $\{\tau_{\min}, \tau_{\min}{+}\Delta\tau, \ldots, \tau_{\max}\}$ using the steepest post-peak drop of the scene-count curve:
\begin{equation}
\Delta N(\tau_i) \;=\; \frac{N(\tau_i{+}\Delta\tau)-N(\tau_i)}{\Delta\tau},
\label{eq:deltaN}
\end{equation}
\begin{equation}
\tau^\star \;=\; \arg\max_{\tau_i}\big(-\,\Delta N(\tau_i)\big).
\label{eq:tau-star}
\end{equation}
\textit{Symbols:} $N(\tau)$—number of detected scenes at threshold $\tau$; $\Delta\tau$—grid step; $\tau^\star$—selected operating threshold.

\textbf{Step 6. Scene Boundary Refinement.} 
To enhance temporal coherence and semantic consistency, we refine the initial scene segmentation by eliminating overly short segments \cite{diao2025temporal, yao2025event}. Such short scenes often arise from noise or subtle visual fluctuations and typically lack sufficient content, leading to \textbf{Scene Boundary Refinement.} To enhance temporal coherence and semantic consistency, we refine the initial scene segmentation by eliminating overly short segments. Such short scenes often arise from noise or subtle visual fluctuations and typically lack sufficient content, leading to unnecessary fragmentation \cite{wu2025cognitive, wu2025image}. We regard a scene as \emph{short} if its length is less than $M = 150$ frames(approximately 5 seconds). In practice, we observe that setting a threshold of 5 seconds helps prevent over-segmentation without removing meaningful events. The initial segmentation stage identifies boundaries primarily through sharp changes in visual features. To further refine these boundaries, we merge back any short segment whose visual representation is highly similar to one of its neighboring segments, ensuring smoother and more consistent temporal units. For a given short scene $S_i$, we first compute its mean embedding:
\begin{equation}
e_i = \frac{1}{|S_i|}\sum_{f \in S_i} \mathbf{f},
\label{eq:mean-embedding}
\end{equation}
where $\mathbf{f}$ denotes the frame-level embedding (e.g., obtained using models such as DINO or CLIP).
Next, we measure the cosine similarity between $e_i$ and the average embeddings of its
two adjacent scenes, $e_{i-1}$ and $e_{i+1}$:
\begin{equation}
\mathrm{sim}_{\mathrm{prev}} =
\frac{e_i \cdot e_{i-1}}{\|e_i\|\|e_{i-1}\|},
\qquad
\mathrm{sim}_{\mathrm{next}} =
\frac{e_i \cdot e_{i+1}}{\|e_i\|\|e_{i+1}\|}.
\label{eq:cosine-sim}
\end{equation}
Finally, the short segment $S_i$ is merged with whichever neighbor produces the higher similarity score, i.e., $\max(\mathrm{sim}_{\mathrm{prev}}, \mathrm{sim}_{\mathrm{next}})$. This refinement step reduces spurious cuts and produces longer, semantically coherent scenes that better align with complete events or actions.
\subsection{Scene description generation}

Produce textual summaries that characterize the content of each scene while also reflecting the narrative of the entire video. After scene boundaries are obtained, we employ a pre-trained video--language model to generate descriptions for every detected scene as well as for the full video. The prompt used for the VideoLLM follows a fixed template,
$\mathcal{P}_{\text{VideoLM}}$: ``Describe this video in detail'',
which directs the model to output detailed and meaningful narratives for the corresponding frame sets.\\
\textbf{To address memory constraints} of VideoLLMs when handling long frame sequences, we adopt a batch-wise procedure: sampled frames are split into manageable batches, the model produces a short description for each batch, and these batch descriptions are then aggregated (concatenated) into a complete narrative for the scene or for the entire video. To preserve coherence across batches, we enforce a consistent formatting rule. Specifically, unless a batch is truly at the beginning or the end of the scene/video, its paragraph is rewritten to start with ``The video continues'' and to conclude with ``The scene concludes,'' rather than ``The video begins/ends.'' This normalization maintains continuity after stitching multiple batch-level descriptions.\\
\textbf{Capturing larger context.} To provide the VideoLLM with a wider temporal window within each batch, we sample frames at $1$ frame per second and select the \emph{middle} frame of each second. This avoids transition or black frames that often appear near cut boundaries and ensures that one representative view is retained from every second. Consequently, the model parses batches that reflect larger temporal spans---and thus broader context---without increasing the input size. In our experiments, this strategy consistently broadened contextual coverage, reduced misinformation, and achieved near-zero omission of important details across videos with varying durations.
This procedure is applied consistently to produce both scene-level descriptions and a holistic description for the full video.\\
\subsection{Scene scores generation}
As shown in the lower-left corner of Figure~\ref{fig:MethodImplementationProcess}, 
this module transforms frame-level annotations into segment-level importance scores based on scene boundaries from the \textit{Scenario Division} stage. 
For the \emph{TVSum} and \emph{SumMe} datasets, \textbf{10\% of the videos} are randomly selected to generate pseudo labels using the averaged frame-level \textit{gtscores}. 
For the \emph{QFVS} dataset, the \textbf{oracle summaries} are directly adopted, where each video is divided into non-overlapping \textbf{5-second shots} to construct corresponding shot-level importance annotations for pseudo-label generation.\\
\textbf{Reason mining and score mapping.}
Before applying reason mining, we map frame-level annotations into segment-level scores, which can be directly used as input for the LLM. The raw annotations are taken from the \textbf{TVSum} and \textbf{SumMe} datasets, both of which provide frame-level \emph{gtscores}. 
Let a video contain $T$ frames, with the annotation score of frame $t$ denoted as $g_t \in [0,1]$ (after normalization). 
Given a prior segmentation, the $i$-th segment is represented by a set of frames $\mathcal{S}_i$. 
The segment-level importance score is defined as the arithmetic mean of the frame scores within that segment:
\begin{equation}
s_i \;=\; \frac{1}{|\mathcal{S}_i|} \sum_{t \in \mathcal{S}_i} g_t ,
\end{equation}

where $|\mathcal{S}_i|$ is the number of frames in the $i$-th segment. 
This value $s_i$ reflects the average importance of the segment within the video. 
We then rank all segments by $s_i$ and select the top three as \emph{high-importance} samples and the bottom three as \emph{low-importance} samples, which form the basis for subsequent reason mining.
\subsection{Scene Scoring Prompt Design}
We guide the LLM to assign scene-level importance scores through carefully engineered prompts grounded in data rather than ad-hoc templates. First, to construct reliable pseudo labels for prompt design, we perform a \emph{reason-mining} step: for each video we select the \textbf{top-3} highest-scoring and \textbf{top-3} lowest-scoring segments (obtained from the GT-derived scene scores), feed their captions into GPT-4o with a strict-JSON prompt, and elicit concise explanations of why the high-score scenes matter, why the low-score scenes do not, and what discriminates the two. We aggregate these rationales into a structured, dataset-adaptive \emph{rubric} that specifies evaluation dimensions and penalties, then instantiate rubric-guided scoring prompts tailored to \textbf{TVSum} (main text) and to \textbf{SumMe}/\textbf{QFVS}. All complete prompt designs (reason-mining prompt, dataset-specific rubric prompts, and contextual scoring templates) are provided in our repository: \textit{ \url{https://github.com/wuyuanli60-svg/Context-Aware-Pseudo-Label-Scoring-for-Zero-Shot-Video-Summarization}}.
\subsubsection{Reason mining with strict JSON prompt.}
Once representative high- and low-score segments are identified, we employ the LLM to perform \emph{reason mining}. 
To ensure balanced and representative supervision, we select the top-$3$ highest-scoring and top-$3$ lowest-scoring segments from each video as reference samples, based on the ground-truth frame-level scores. The textual descriptions of these selected segments—derived from the scene division stage—are fed into GPT-4o with a structured prompt 
that enforces strict JSON-format outputs. The model is required to summarize the distinguishing characteristics between high- and low-importance scenes and articulate the underlying reasons that contribute to their scores. These extracted rationales are then distilled into dataset-specific rubric blocks, which serve as the foundation for constructing our final scoring prompts, while the core part is shown below:
\begin{quote}
\small
\ttfamily
You will write THREE concrete reasons for this video and return STRICT JSON with the keys:\\[3pt]
-- "reason\_positive": one succinct but specific reason why the HIGH-score segments are key.\\
-- "reason\_negative": one succinct but specific reason why the LOW-score segments are not key.\\
-- "reason\_difference": one succinct but specific reason explaining their essential difference.
\end{quote}
This strict JSON format improves interpretability by explicitly grounding the reasons in observable visual elements and actions. At the same time, it provides a transparent foundation for deriving dataset-specific scoring rubrics in subsequent steps.

\subsubsection{From reasons to dataset-specific rubrics.}
After collecting reasons for high- and low-score segments, we move beyond manual inspection by introducing a \emph{second-stage abstraction prompt} that compels the LLM to consolidate these cues into a hierarchical, dataset-aware rationale schema. Concretely, the LLM is instructed to: (i) cluster recurring positive/negative/difference cues across videos; (ii) elevate them into \emph{evaluation dimensions} with explicit constraints and penalties; (iii) add category-specific checklists capturing dataset idiosyncrasies; and (iv) formalize a calibration ladder and an exact $[0,100]$ output rule. The result is a multi-level, interpretable rationale chain—\emph{evidence} $\rightarrow$ \emph{cue} $\rightarrow$ \emph{dimension} $\rightarrow$ \emph{scoring rule}—that preserves dataset characteristics while remaining generalizable. This structured rationale is then compiled into rubric-guided scoring prompts used by \textbf{GPT-5} at inference, enabling consistent, bias-resistant scene scoring and reducing sensitivity to prompt wording.

\textbf{TVSum rubric prompt (main text).}
Below shows the main scoring rubric for TVSum:
\begin{quote}
\small
\ttfamily
1) Task/Thematic Relevance (0–35)\\
   - Advances the core activity (hands-on steps:\\ grooming/feeding/repair/cooking; riders in motion; performers/floats).\\
   - Low if generic or unrelated.\\[4pt]
2) Action/Interaction \& Skill (0–20)\\
   - Concrete interactions (human–animal, human–tool, performer–crowd, rider–bike), visible skill/risk.\\[4pt]
3) Detail \& Visibility (0–15)\\
   - Close-ups of hands/tools/animals/ingredients/parts; clarity to learn the step; hygiene/safety cues add credibility.\\[4pt]
4) Informational Uniqueness (0–15)\\
   - New step/angle/outcome vs. adjacent scenes; penalize near-duplicates.\\[4pt]
5) Narrative Progression (0–15)\\
   - Bridges setup→action→result or marks a turning point/outcome verification.\\[4pt]
Penalties: title/logo/blank (-15), off-topic (-10), static (-8), low visibility (-6), redundancy (-6).\\
Final score = round(0.35R + 0.20A + 0.15D + 0.15U + 0.15N + PrefAdj), clamp [0,100].\\
Output exactly one integer in [0,100].
\end{quote}

\subsubsection{Score generation with contextualized inputs}
We use two prompts. Below are the operational instructions actually fed to the LLM.\\
\textbf{Boundary scenes (first/last) — Target-only prompt.}
\begin{quote}
\small
\ttfamily
1) Use the dataset-specific rubric (TVSum or SumMe).\\
2) Inputs: (a) Global video description; (b) Target scene description.\\
3) No local context: ignore previous/next scenes entirely.\\
4) Score ONLY the target scene per rubric dimensions and penalties.\\
5) If a user preference is provided, apply a small modifier to the Relevance dimension only.\\
6) Output EXACTLY ONE integer in 0--100 (no words, no units).
\end{quote}
\textbf{Intermediate scenes — Contextualized prompt.}
\begin{quote}
\small
\ttfamily
1) Use the dataset-specific rubric (TVSum or SumMe).\\
2) Inputs: (a) Target scene (score this); (b) Global video description; \\(c) Previous scene (context only); (d) Next scene (context only).\\
3) Internally refine Previous and Next into 1--2 short notes each: who/what action, \\stage (setup/key/aftermath), new vs.\ repeated info, visibility; do NOT reveal these notes.\\
4) Base score comes PRIMARILY from the Target + Global per rubric; neighbors are only for a small adjustment.\\
5) Apply a conservative context adjustment (\(\pm 5\)): +5 if the Target clearly adds NEW \\information/progression vs.\ \emph{both} neighbors; \(-5\) if largely DUPLICATED vs.\ \emph{both}; 0 if unclear/mixed.\\
6) If a user preference is provided, apply ONLY a subtle modifier to Relevance (at most \(\pm 5\)); do not alter other dimensions.\\
7) Always SCORE ONLY THE TARGET; neighbors are reference signals, not items to be scored.\\
8) Output EXACTLY ONE integer in 0--100 (no words, no units).
\end{quote}

\subsection{Frame-level scoring}
Produce fine-grained importance values for each frame, so that every moment in the video is assessed according to both its own content and its temporal context.We adopt a three-stage pipeline. First, each frame is initialized by inheriting the score of the scene it belongs to. Second, a temporal smoothing function is applied around scene boundaries to avoid abrupt discontinuities and to ensure gradual score transitions between scenes. Finally, a representativeness weight is assigned to each frame inside its scene, highlighting the most informative or typical frames. The final per-frame score is obtained by combining the smoothed scene score with the frame-level weight.\\
\textbf{Scene to frames: normalization and temporal smoothing.}
After obtaining scene-level scores, we map them to frame-level values. 
First, all scene scores are normalized to a common scale to enhance relative contrast. 
Let $s_i$ be the raw score of scene $i$ and $\mathcal{S}_i$ the set of its frames. 
We write
\begin{equation}
\tilde{s}_i = \mathrm{Norm}(s_i).
\end{equation}
Each frame initially inherits its parent-scene value:
\begin{equation}
z^{(0)}_{t} = \tilde{s}_i, \qquad t \in \mathcal{S}_i .
\end{equation}

To avoid discontinuities at scene boundaries, we interpolate between consecutive scenes using a cosine schedule defined on the interval between their midpoints. 
Let $m_i$ and $m_{i+1}$ be the midpoints of scenes $i$ and $i{+}1$. 
For frames $t \in [m_i,\, m_{i+1}]$, define the interpolation weight
\begin{equation}
\alpha_t \;=\; \frac{1 - \cos\!\left(\pi\,\frac{t - m_i}{\,m_{i+1} - m_i\,}\right)}{2}.
\end{equation}
The smoothed frame value is then
\begin{equation}
z^{(1)}_{t} \;=\; (1-\alpha_t)\,\tilde{s}_i \;+\; \alpha_t\,\tilde{s}_{i+1}.
\end{equation}
For frames outside $[m_i,\, m_{i+1}]$, we simply keep $z^{(1)}_{t}=\tilde{s}_i$. 
This cosine-based blending yields gradual transitions across scene boundaries while preserving the relative importance imposed by the normalized scene scores.\\
\textbf{Frame weighting within a scene.}
After smoothing scene-level scores, we further assign a representativeness weight to each frame within its scene. 
The intuition is to highlight frames that are either visually stable (high internal consistency) or contain distinct changes (high uniqueness). 
To this end, each scene is divided into non-overlapping segments of length $W$, and two complementary measures are computed for each segment $S$: consistency and uniqueness.

Consistency is obtained by clustering all frame embeddings of the scene with $K$-means, where the optimal number of clusters $K^\ast$ is determined via the elbow method on the WCSS curve. 
Given a sequence of cluster labels $\{l_{t+1},\ldots,l_{t+W}\}$ for frames in $S$, the consistency score is defined as
\begin{equation}
\mathrm{Consistency}(S) \;=\; \frac{1}{W}\max_{c}\sum_{i=1}^{W}\mathbf{1}\!\left[l_{t+i}=c\right].
\end{equation}

Uniqueness is defined as the average $\ell_2$ distance between each frame embedding and the mean embedding of the segment. 
Let $\{\mathbf{f}_{t+1},\ldots,\mathbf{f}_{t+W}\}$ denote the embeddings and $\bar{\mathbf{f}}$ their mean, 
\begin{equation}
\bar{\mathbf{f}} \;=\; \frac{1}{W}\sum_{i=1}^{W}\mathbf{f}_{t+i},
\end{equation}
\begin{equation}
\mathrm{Uniqueness}(S) \;=\; \frac{1}{W}\sum_{i=1}^{W}\left\lVert \mathbf{f}_{t+i}-\bar{\mathbf{f}}\right\rVert_2.
\end{equation}

The representativeness weight of segment $S$ is then a convex combination of consistency and uniqueness, controlled by a mixing parameter $\sigma \in [0,1]$:
\begin{equation}
\mathrm{Weight}(S) \;=\; \sigma \cdot \mathrm{Consistency}(S) \;+\; (1-\sigma)\cdot \mathrm{Uniqueness}(S).
\end{equation}
This value is shared by all frames in $S$.

To adapt across different video lengths $T$, both $\sigma$ and $W$ are chosen according to predefined rules based on a short-video threshold $S$ (set to 1.8 minutes, approximately 100 seconds):
\begin{equation}
F_S(\mathrm{Video})=\begin{cases}
\{\sigma=0.1,\; W=1\ \mathrm{s}\}, & T>5S,\\[6pt]
\{\sigma=1.0,\; W=1\ \mathrm{s}\}, & S<T\le 5S,\\[6pt]
\{\sigma=0.3,\; W=3\ \mathrm{s}\}, & \text{otherwise}.
\end{cases}
\end{equation}

In practice, long videos emphasize novelty through smaller $\sigma$ and finer $W$, medium-length videos favor stability with larger $\sigma$, while short videos employ longer segments to improve robustness.

\section{Experiments and Results}

In this section, we present a comprehensive set of experiments to evaluate our method against state-of-the-art approaches. 
We consider both conventional video summarization benchmarks and text-guided scenarios. The section begins with results on standard benchmarks and is followed by a component-wise ablation study of our framework.

\subsection{Standard Video Summarization}

\textbf{Datasets.}
We evaluate our approach on two widely recognized datasets: \textbf{SumMe} and \textbf{TVSum}. 
The \emph{SumMe} dataset contains 25 short videos with durations between 1 and 6 minutes. 
The content spans diverse contexts such as holidays, sports, and social events, recorded from both egocentric and third-person viewpoints. 
Each video is annotated by 15--18 individuals, who select key fragments. 
As a result, every video is paired with multiple human-generated summaries of varying lengths, typically covering 5\%--15\% of the original duration. 
The \emph{TVSum} dataset consists of 50 videos, ranging in length from 1 to 61 minutes, grouped into 10 activity categories (e.g., repairing a tire, preparing food, participating in parades). 
Each category includes 5 videos, collected from YouTube. 
Annotations were obtained from 20 users who assigned frame-level importance scores on a discrete scale from 1 (least important) to 5 (most important).\\
\textbf{Baselines.}
We evaluate our approach on two standard video summarization datasets, \textbf{SumMe} and \textbf{TVSum}, and compare it against a variety of existing methods. The baselines include \textbf{supervised methods} such as re-SEQ2SEQ~\cite{reseq2seq}, CLIP-It~\cite{clipit}, MAVS~\cite{mavs}, iPTNet~\cite{iptnet}, A2Summ~\cite{a2summ}, and PGL-SUM~\cite{pglsum}, which rely on annotated training data; \textbf{unsupervised methods} such as AC-SUM-GAN~\cite{acsumgan}, RS-SUM~\cite{rssum}, and SegSum~\cite{segsum}, which generate summaries without requiring manual labels; and recent \textbf{zero-shot methods}, in particular \textbf{PROMPTS TO SUMMARIES: ZERO-SHOT LANGUAGE-GUIDED VIDEO SUMMARIZATION}~\cite{prompts2summaries}, which leverages large language models and prompt engineering to produce summaries without additional training. Our framework is mainly compared with \textbf{PROMPTS TO SUMMARIES: ZERO-SHOT LANGUAGE-GUIDED VIDEO SUMMARIZATION} as the primary baseline. We conduct evaluations on the \textbf{TVSum} and \textbf{SumMe} datasets in terms of \textbf{precision (P)}, \textbf{recall (R)}, and \textbf{F-score}, and further provide visual comparisons to intuitively demonstrate the differences between our method and this baseline.\\
\textbf{Evaluation Protocol.} We employ the keyshot-based evaluation metric, which measures the agreement between an automatically generated summary $A$ and a user reference summary $B$. Both $A$ and $B$ are represented by sets of selected frames. 
The evaluation first computes precision and recall in terms of temporal overlap:
\begin{equation}
P = \frac{|A \cap B|}{|A|}, 
\qquad 
R = \frac{|A \cap B|}{|B|},
\end{equation}
where $|A|$ and $|B|$ denote the lengths (in frames) of the generated and reference summaries, respectively. 
The F1-score is then defined as
\begin{equation}
F = 2 \cdot \frac{P \cdot R}{P + R}.
\end{equation}

For \textbf{SumMe}, user annotations are provided directly as key fragments, allowing straightforward comparison with system outputs. 
For \textbf{TVSum}, the annotations consist of frame-level importance scores. 
Following the protocol in TVSum~\cite{tvsum}, we first transform frame-level annotations into key fragments before applying the F1 computation. 
The aggregation strategy differs across the two datasets: on TVSum, the F1 for a given video is averaged across all users, while on SumMe the maximum F1 among users is reported, consistent with prior work~\cite{summe}. 

Finally, to obtain dataset-level performance, we compute the mean F1-score over five independent splits of the data into training and testing sets, in line with the standard practice established in~\cite{keval}. 
This average serves as the principal metric for quantitative evaluation.\\
\textbf{Design Choices.} In the normalization stage, three options are considered: 
\emph{Min-Max normalization}, \emph{Exponential normalization}, and a mixed strategy. 
The choice of method depends on the intrinsic characteristics of each dataset.\\
\textbf{Normalization for SumMe.} 
For the SumMe dataset, we adopt Min-Max normalization, which linearly maps values into the range [0,1]. 
This scheme retains proportional differences while remaining sensitive to outliers, making it well-suited to the variability caused by subjective user preferences in SumMe. 
As highlighted in~\cite{gygli2014summe}, the evaluation of SumMe relies on the maximum F1-score across all references to select the closest user summary. 
In this context, Min-Max normalization provides a better match between computed importance scores and the diversity of human annotations.\\
\textbf{Normalization for TVSum.}
For the TVSum dataset, we apply Exponential normalization. 
Unlike Min-Max scaling, this transformation amplifies higher values and therefore accentuates more salient segments. 
Such a property is particularly appropriate for TVSum, where user annotations tend to be consistent and uniformly distributed. 
Since TVSum evaluation is based on the average F1-score across all annotators, Exponential normalization ensures better alignment with the relatively stable annotation patterns observed in this dataset. 
\begin{table}[b]
\centering
\caption{Performance evaluation on TVSum and SumMe (Precision, Recall, F1).}
\label{tab:results}
\begin{tabular}{lcc}
\toprule
\textbf{Metric} & \textbf{SumMe} & \textbf{TVSum} \\
\midrule
Precision & 56.62& 62.59\\
Recall    & 58.81& 63.16\\
F1-Score  & 57.58& 63.05\\
\bottomrule
\end{tabular}
\end{table}
\vspace{-0.2cm}
\subsection{Standard Video Summarization Evaluation }
\begin{table*}[t]
\centering
\small
\setlength{\tabcolsep}{6pt} 
\caption{Comparison on SumMe~\cite{summe} and TVSum~\cite{tvsum}. 
The proposed zero-shot model yields a notable gain over the previous zero-shot baseline, outperforms all unsupervised methods, 
and approaches the results of supervised models. The results cover methods published between 2016 and 2025, reflecting 
the evolution from supervised learning to zero-shot prompting.}
\begin{tabular}{l c c r r}
\toprule
\makecell[l]{\textbf{Method name}} 
& \makecell[c]{\textbf{Year}} 
& \makecell[c]{\textbf{Doesn’t require}\\\textbf{training data}}
& \textbf{SumMe} 
& \textbf{TVSum} \\
\midrule

\multicolumn{5}{l}{\textbf{Supervised methods}}\\
re-SEQ2SEQ~\cite{reseq2seq}   & 2018 & \xmark & --   & 63.9  \\
CLIP-It~\cite{clipit}         & 2021 & \xmark & 54.2 & 66.3  \\
MAVS~\cite{mavs}              & 2023 & \xmark & --   & 67.5  \\
iPTNet~\cite{iptnet}          & 2022 & \xmark & 54.5 & 63.4  \\
A2Summ~\cite{a2summ}          & 2021 & \xmark & 55.0 & 63.4  \\
PGL-SUM~\cite{pglsum}         & 2018 & \xmark & 57.1 & --    \\[2pt]
\midrule

\multicolumn{5}{l}{\textbf{Unsupervised methods}}\\
AC-SUM-GAN~\cite{acsumgan}    & 2020 & \xmark & 50.8  & 60.6  \\
RS-SUM~\cite{rssum}           & 2019 & \xmark & 52.0  & 61.4  \\
SegSum~\cite{segsum}          & 2016 & \xmark & 54.07 & 62.02 \\[2pt]
\midrule

\multicolumn{5}{l}{\textbf{Zero-shot methods}}\\
PROMPTS TO SUMMARIES~\cite{prompts2summaries} 
                              & 2025 & \cmark & \textbf{56.73} & \textbf{62.21} \\
Ours                          & 2025 & \cmark & \textbf{57.58} & \textbf{63.05} \\

\bottomrule

\end{tabular}

\label{tab:zero_shot_summary}

\end{table*}
\begin{table*}[t]
\centering
\small
\setlength{\tabcolsep}{6pt}
\caption{
Performance on QFVS~\cite{sharghi2017qfvs}. 
Baselines are supervised and require training data, while our method is zero-shot and training-free.
\textbf{Bold} values denote the highest F1 score per column, and \underline{underlined} values indicate scores higher than those of \textit{PROMPTS TO SUMMARIES}.
}
\begin{tabular}{l c c r r r r r}
\toprule
\makecell[l]{\textbf{Method name}} &
\makecell[c]{\textbf{Year}} &
\makecell[c]{\textbf{Doesn’t require}\\\textbf{training data}} &
\textbf{Vid 1} & \textbf{Vid 2} & \textbf{Vid 3} & \textbf{Vid 4} & \textbf{Avg} \\
\midrule

\multicolumn{8}{l}{\textbf{Supervised methods}}\\
SeqDPP~\cite{gong2014seqdpp}           & 2014 & \xmark & 36.59 & 43.67 & 25.26 & 18.15 & 30.92 \\
SH-DPP~\cite{sharghi2016qfes}          & 2016 & \xmark & 35.67 & 42.74 & 36.51 & 18.62 & 33.38 \\
QFVS~\cite{sharghi2017qfvs}            & 2017 & \xmark & 48.68 & 41.66 & 56.47 & 29.96 & 44.19 \\
CLIP-It: ResNet~\cite{clipit}          & 2021 & \xmark & 55.19 & 51.03 & 64.26 & 39.47 & 52.49 \\
CLIP-It: CLIP-Image~\cite{clipit}      & 2021 & \xmark & \textbf{57.13} & 53.60 & \textbf{66.08} & 41.41 & \textbf{54.44} \\
\midrule

\multicolumn{8}{l}{\textbf{Zero-shot methods}}\\
PROMPTS TO SUMMARIES~\cite{prompts2summaries} & 2025 & \cmark & 53.57 & \textbf{55.66} & 63.25 & 41.20 & 53.42 \\
\textbf{Ours}                                 & 2025 & \cmark & \underline{54.06}& 54.32 & \underline{64.22}& \underline{\textbf{42.56}} & \underline{53.79}\\

\bottomrule
\end{tabular}
\label{tab:qfvs_cmp}
\end{table*}
\textbf{Precision, Recall, and F1 Score.} 
We evaluate our model on the SumMe and TVSum datasets using precision, recall, and F1 score to measure summarization quality under a zero-shot setting. 
As reported in Table~\ref{tab:results}, the proposed model achieves balanced precision and recall across both datasets, yielding final F1-scores of \textbf{57.58\%} on SumMe and \textbf{63.05\%} on TVSum. 
The close alignment among precision, recall, and F1 suggests that the model consistently identifies relevant segments while maintaining good coverage of the ground-truth annotations. \\
\textbf{Performance comparison.}
As shown in Table~\ref{tab:zero_shot_summary}, our zero-shot model achieves \textbf{57.58\%} on the SumMe dataset and \textbf{63.05\%} on the TVSum dataset. 
These results demonstrate clear and consistent improvements across both benchmarks. 
Compared to the existing zero-shot baseline (PROMPTS TO SUMMARIES~\cite{prompts2summaries}), our model yields a gain of \textbf{+0.85\%} on SumMe and \textbf{+0.84\%} on TVSum, 
highlighting the effectiveness of our rubric-guided reasoning and contextual scoring design.
Beyond zero-shot methods, our approach also \textbf{outperforms all unsupervised baselines}, including AC-SUM-GAN~\cite{acsumgan}, RS-SUM~\cite{rssum}, and SegSum~\cite{segsum}. 
Specifically, it surpasses the best-performing unsupervised method (SegSum) by \textbf{+3.51\%} on SumMe and \textbf{+1.03\%} on TVSum. 
Remarkably, the proposed zero-shot framework even approaches the results of several supervised models such as A2Summ~\cite{a2summ} and iPTNet~\cite{iptnet}, 
while requiring \emph{no training data, annotations, or dataset-specific adaptation}. 

Overall, these findings indicate that our prompt-based zero-shot summarization framework not only closes the gap between unsupervised and supervised paradigms, 
but also establishes a new performance level within the zero-shot regime, achieving competitive results on both datasets without any gradient-based optimization.


\subsection{Query-Focused Video Summarization}
\textbf{Datasets.}
We evaluate our query-focused video summarization approach on the \textbf{QFVS} benchmark. \textbf{QFVS} is built on the \textbf{UT Egocentric (UTE)} dataset and comprises four long videos (about 3--5 hours each). Every video is paired with roughly 45--46 user queries and human-annotated summaries~\cite{sharghi2017qfvs,lee2012ute}. Each video is segmented into non-overlapping 5-second shots and annotated with concise scene captions (e.g., \textit{SKY}, \textit{STREET}, \textit{ROAD}, \textit{CAR}). Annotations from three raters are aggregated at the video--query level into a caption-dense representation, from which query-conditioned summaries are produced via a greedy procedure. The resulting aggregated ``\textbf{Oracle}'' summaries show higher agreement with the querying user than the generic inter-annotator consensus~\cite{sharghi2017qfvs}.\\
\textbf{Evaluation Protocol.}
Although prior work typically predicts a single score per shot, our model produces frame-level importance scores. Following the evaluation methodology of CLIP-It~\cite{clipit}, we convert frame-level predictions into shot-level scores by averaging all frame scores within each shot. For the \textbf{QFVS} dataset, since each video is pre-segmented into non-overlapping 5-second shots~\cite{sharghi2017qfvs}, our evaluation is accordingly performed at the shot level. The final video summary is generated by selecting the shots with the highest predicted scores, with the total duration constrained to match the length of the video–query Oracle summary, as described in~\cite{sharghi2017qfvs}.\\
\textbf{Quantitative analysis.}
As shown in Table~\ref{tab:qfvs_cmp}, our zero-shot framework attains F1 scores of 54.06 (Vid1), 54.32 (Vid2), 64.22 (Vid3), and 42.56 (Vid4), with an overall average of \textbf{53.79}.Compared with the zero-shot baseline \emph{PROMPTS TO SUMMARIES}~\cite{prompts2summaries} (53.57, 55.66, 63.25, 41.20; average 53.42), our method improves Vid1 by 0.49 points, Vid3 by 0.97 points, and Vid4 by 1.36 points, and raises the average by 0.37 points; performance on Vid2 is lower by 1.34 points.
The isolated decrease on Vid2 is plausibly due to occasional tension between the user query and the pseudo reasoning/chain-of-thought introduced by our framework—both provide strong guidance, and when their emphases diverge, the internal reasoning may bias selection away from the query’s most salient evidence.

Overall, the consistent gains on three videos and the higher average indicate that integrating pseudo labels and query-conditioned context enhances query alignment and temporal coherence while remaining training-free, thereby narrowing the gap to supervised alternatives.
\vspace{-0.15cm}
\begin{table*}[!t]
  \centering
  \caption{Ablation on TVSum and SumMe. Higher is better (↑). Best per column is \textbf{bold}.}
  \label{tab:ablation_tvsum_summe}
  \setlength{\tabcolsep}{8pt}
  \begin{tabular}{lccc ccc}
    \toprule
    \multirow{2}{*}{Method} & \multicolumn{3}{c}{TVSum} & \multicolumn{3}{c}{SumMe} \\
    \cmidrule(lr){2-4}\cmidrule(lr){5-7}
     & F1 $\uparrow$ & Precision $\uparrow$ & Recall $\uparrow$
     & F1 $\uparrow$ & Precision $\uparrow$ & Recall $\uparrow$ \\
    \midrule
    Baseline~\cite{prompts2summaries}& 62.21 & 62.22 & 62.22 & 56.73 & 55.77 & 57.96 \\
    +Pseudo Label       & 62.83 & 62.53 & 63.01 & 57.24 & 56.31 & 58.61 \\
    +Pseudo +Context    & \textbf{63.05} & \textbf{62.79} & \textbf{63.16}
                         & \textbf{57.58} & \textbf{56.62} & \textbf{58.81} \\
    \bottomrule
  \end{tabular}
\end{table*}
\begin{table*}[t]
\centering
\small
\setlength{\tabcolsep}{6pt}
\caption{Ablation study on QFVS~\cite{sharghi2017qfvs}. Higher is better (↑). 
Best per column is in \textbf{bold}. For non-baseline rows, each value is annotated with its change relative to the Baseline (↑/↓ with absolute difference).}
\begin{tabular}{l c c c c c}
\toprule
\textbf{Method} & \textbf{Vid 1 (F1)} & \textbf{Vid 2 (F1)} & \textbf{Vid 3 (F1)} & \textbf{Vid 4 (F1)} & \textbf{Avg (F1)} \\
\midrule
Baseline~\cite{prompts2summaries}       
& 53.57 & 55.66 & 63.25 & 41.20 & 53.42 \\
\midrule
+ Pseudo Label                          
& 54.78 {\scriptsize(↑\,1.21)} 
& 42.54 {\scriptsize(↓\,13.12)} 
& 63.46 {\scriptsize(↑\,0.21)} 
& 43.41 {\scriptsize(↑\,2.21)} 
& 51.04 {\scriptsize(↓\,2.38)} \\
\textbf{+ Pseudo + Context (Ours)}      
& \textbf{54.06} {\scriptsize(↑\,0.49)} 
& \textbf{54.32} {\scriptsize(↓\,1.34)} 
& \textbf{64.22} {\scriptsize(↑\,0.97)} 
& \textbf{42.56} {\scriptsize(↑\,1.36)} 
& \textbf{53.79} {\scriptsize(↑\,0.37)} \\
\bottomrule
\end{tabular}
\label{tab:qfvs_ablation}
\end{table*}
\subsection{Ablation Study}
This study investigates two components of our zero-shot framework: (i) \emph{pseudo-label reconstruction prompts} and (ii) \emph{contextualized scoring}, where the reconstructed prompts are augmented with local scene context (summaries of the preceding and following segments). All variants use identical data splits and evaluation protocols so that performance differences arise solely from prompt design. The quantitative results on SumMe/TVSum (Table~\ref{tab:ablation_tvsum_summe}) and the QFVS ablations (Table~\ref{tab:qfvs_ablation}) jointly substantiate the effectiveness of the proposed framework.

\textbf{Effects on SumMe/TVSum (Table~\ref{tab:ablation_tvsum_summe}).}
Relative to the plain zero-shot baseline, introducing pseudo-label reconstruction increases F1 from 56.73 to 57.24 on \textbf{SumMe} (+0.51) and from 62.21 to 62.83 on \textbf{TVSum} (+0.62), accompanied by consistent gains in precision and recall. Adding contextualized scoring on top of pseudo labels further raises F1 to 57.58 on SumMe (+0.34 over pseudo labels; +0.85 over baseline) and to 63.05 on TVSum (+0.22 over pseudo labels; +0.84 over baseline). These results indicate that short summaries of adjacent scenes enhance temporal coherence and reduce redundancy, yielding more consistent frame-importance distributions.

\textbf{Effects on QFVS (Table~\ref{tab:qfvs_ablation})}
Using \emph{PROMPTS TO SUMMARIES} as the zero-shot baseline, adding \emph{pseudo-label reconstruction} alone yields 54.78, 42.54, 63.46, and 43.41 (Avg=51.04), i.e., relative changes of +1.21 (Vid1), $-13.12$ (Vid2), +0.21 (Vid3), and +2.21 (Vid4), with an average decrease of $-2.38$. This pattern shows that while pseudo supervision can help, a mismatch between the strongly guided pseudo reasoning/chain-of-thought and the user query may occasionally bias selection away from the query’s most salient evidence (notably on Vid2). When \emph{contextualized scoring} is added on top of pseudo labels, the scores become 54.06, 54.32, 64.22, and 42.56 (Avg=53.79), corresponding to +0.49 (Vid1), $-1.34$ (Vid2), +0.97 (Vid3), +1.36 (Vid4), and an average improvement of +0.37 over the baseline. Moreover, stability improves markedly: the $L_1$ sum of absolute deviations from the baseline across the four videos drops from 16.75 (pseudo only) to 4.16 (pseudo + context), and the large negative deviation on Vid2 contracts from $-13.12$ to $-1.34$. These findings confirm that contextual cues effectively mediate the potential tension between query guidance and internal reasoning, leading to more stable and reliable outcomes across all validation videos.

Across SumMe, TVSum, and QFVS, pseudo-label reconstruction systematically improves alignment with salient events and human annotation tendencies; augmenting with contextualized scoring further enhances temporal coherence and robustness, delivering the best overall accuracy and stability while preserving the training-free, zero-shot setting.

\definecolor{myyellow}{HTML}{DAA520}
\begin{figure*}[!ht]
  \centering
  \setlength{\tabcolsep}{2pt}
  \renewcommand{\arraystretch}{1.02}

  \newcommand{\rowimg}[1]{%
    \includegraphics[width=0.95\linewidth,keepaspectratio,clip,trim=10 10 12 12]{\detokenize{#1}}%
  }

  \begin{tabular}
  {@{} >{\centering\arraybackslash}m{0.14\textwidth}   
                    >{\centering\arraybackslash}m{0.21\textwidth}   
                    >{\centering\arraybackslash}m{0.21\textwidth}
                    >{\centering\arraybackslash}m{0.21\textwidth}
                    >{\centering\arraybackslash}m{0.21\textwidth} @{}}

    \textbf{Pipeline Stage} &
    \textbf{Bus-in-Rock-Tunnel (SumMe)} &
    \textbf{Eiffel-Tower (SumMe)} &
    \textbf{qqR6AEXwxoQ (TVSum)} &
    \textbf{vdmoEJ5YbrQ (TVSum)} \\

    \makecell[c]{\textbf{(a) Initial Scene}\\[-2pt]
      \textbf{boundaries}\\
      \textcolor{blue}{\footnotesize User annotations (mean)}\\[-2pt]
      \textcolor{red}{\footnotesize Scene boundaries}}
    &
    \rowimg{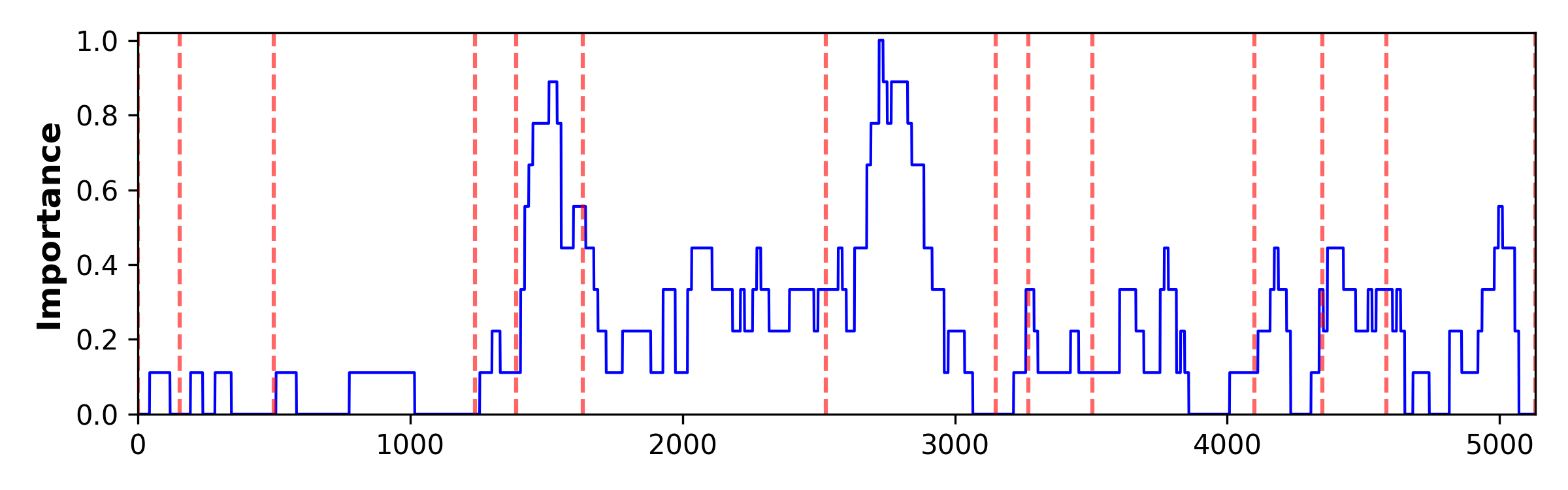} &
    \rowimg{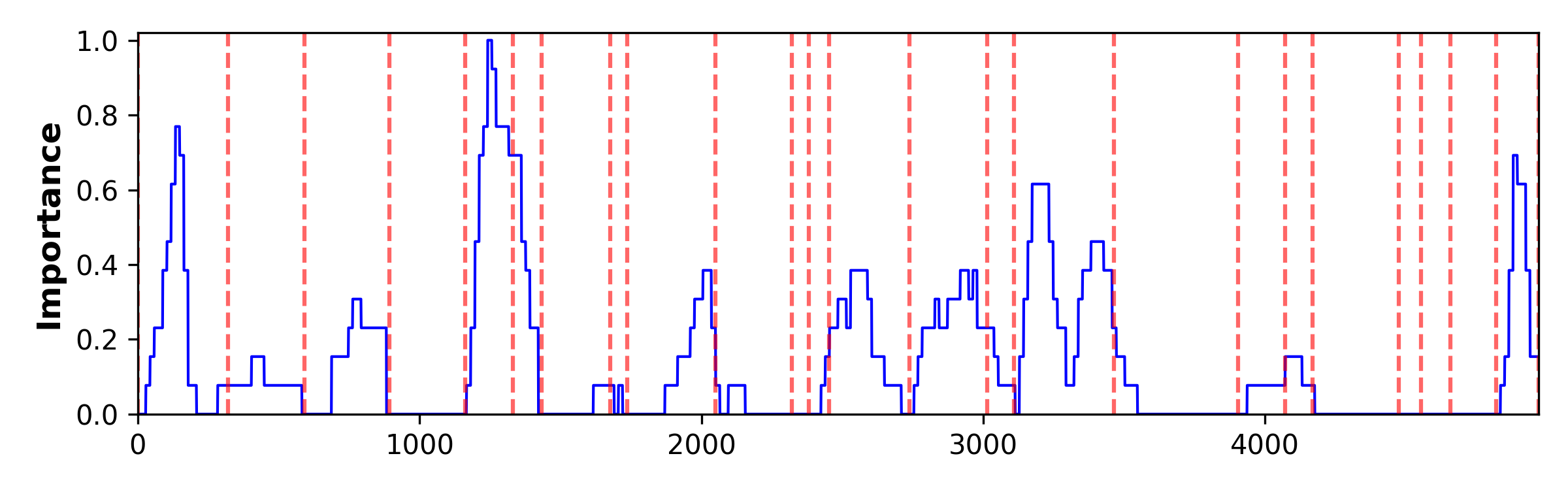} &
    \rowimg{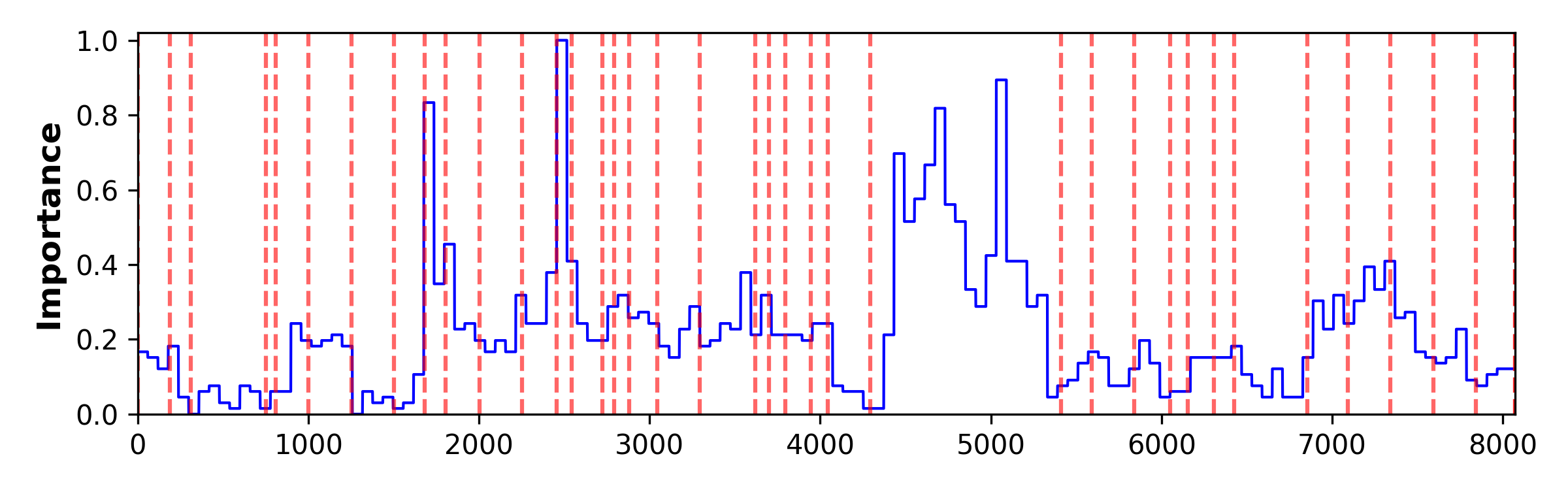} &
    \rowimg{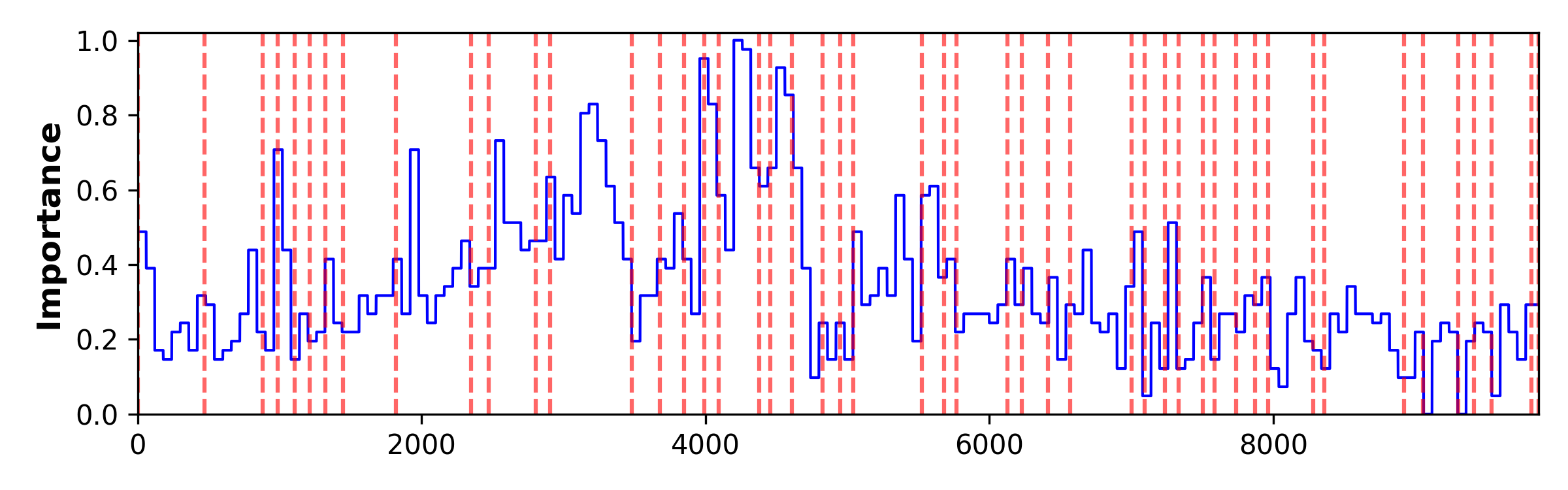} \\[-2pt]

    \makecell[c]{\textbf{(b) Refined Scene}\\[-2pt]
      \textbf{boundaries}\\
      \textcolor{blue}{\footnotesize User annotations (mean)}\\[-2pt]
      \textcolor{red}{\footnotesize Scene boundaries}}
    &
    \rowimg{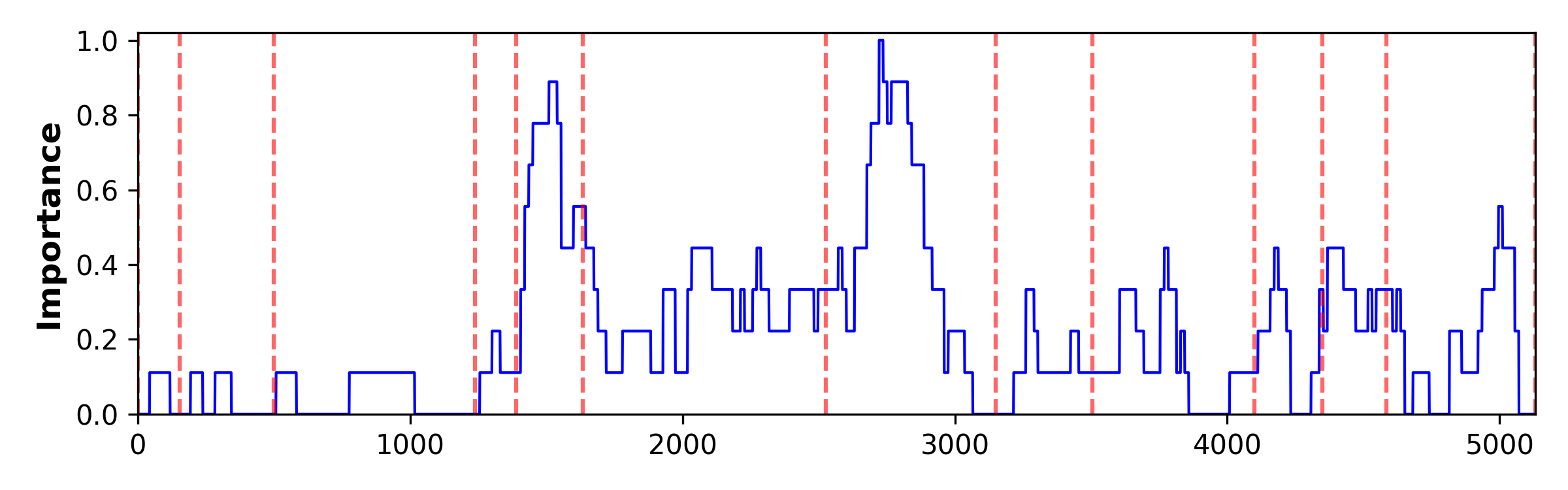} &
    \rowimg{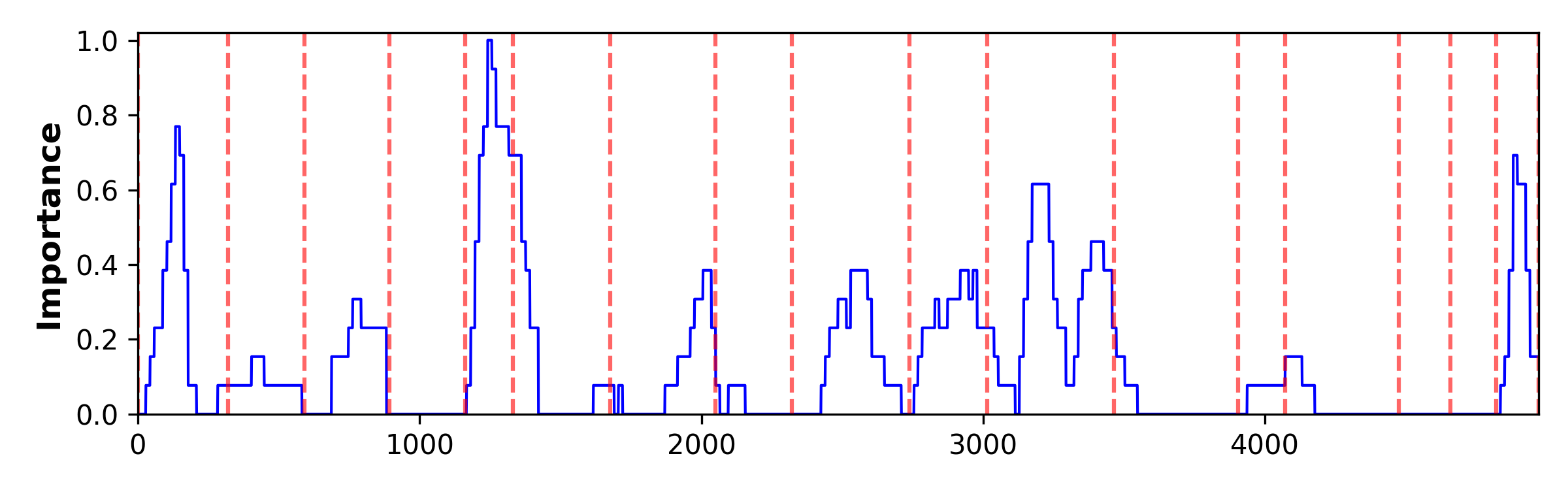} &
    \rowimg{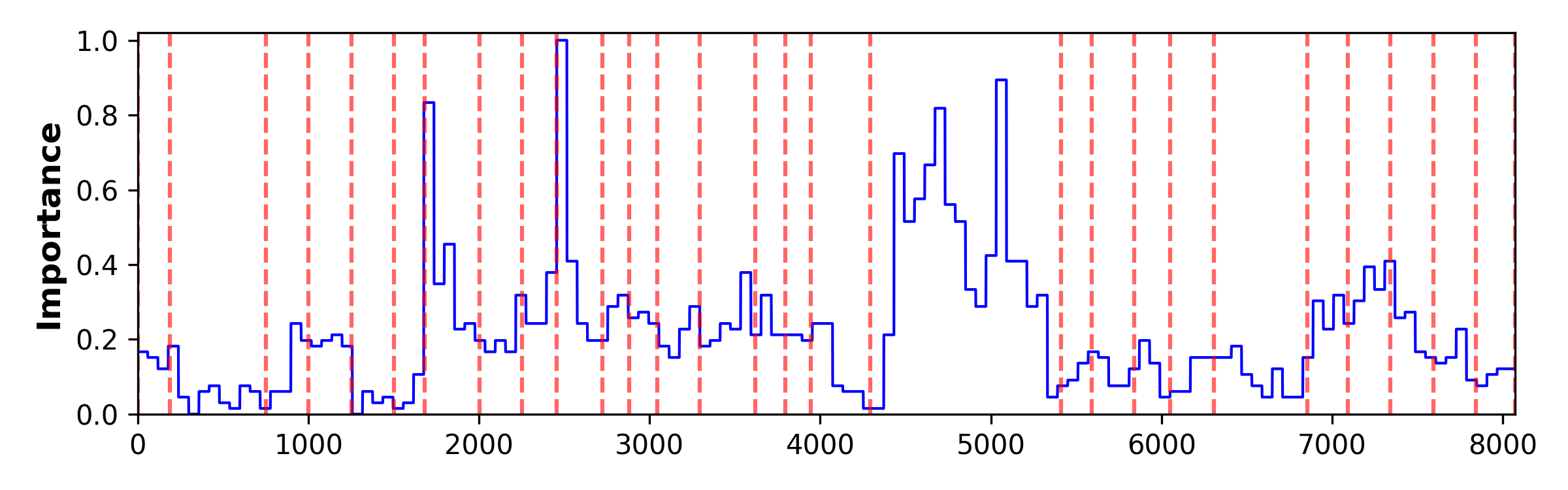} &
    \rowimg{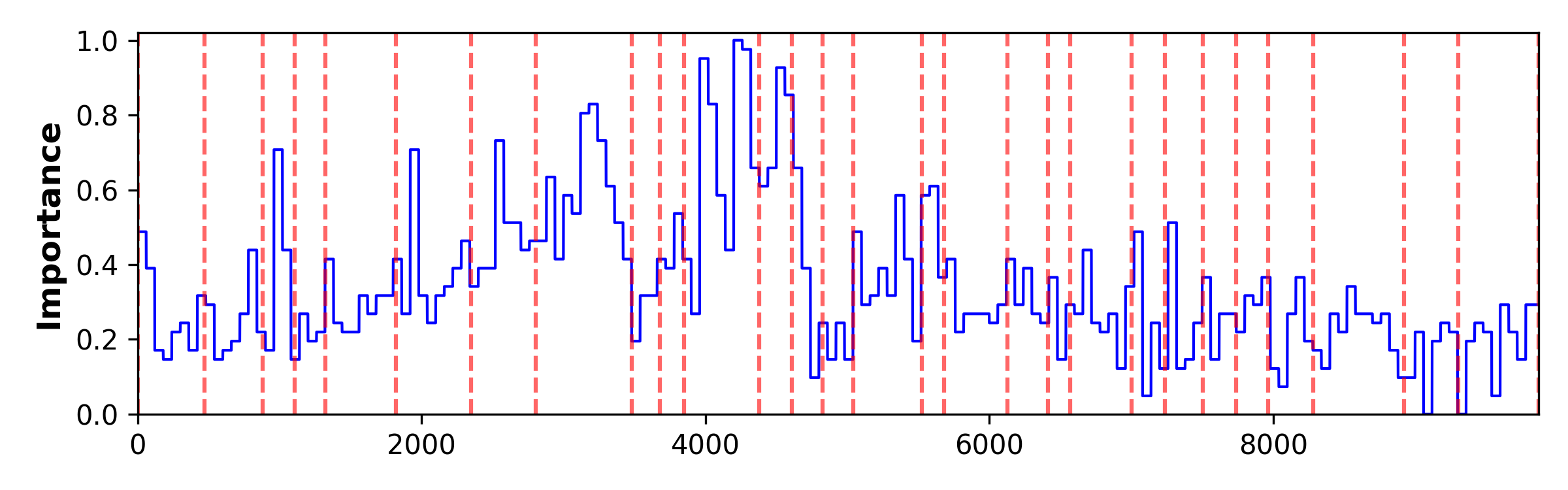} \\[-2pt]

    \makecell[c]{\textbf{(c) Scene Scores}\\
      \textcolor{blue}{\footnotesize User annotations (mean)}\\[-2pt]
      \textcolor{myyellow}{\footnotesize Frame scores}}
    &
    \rowimg{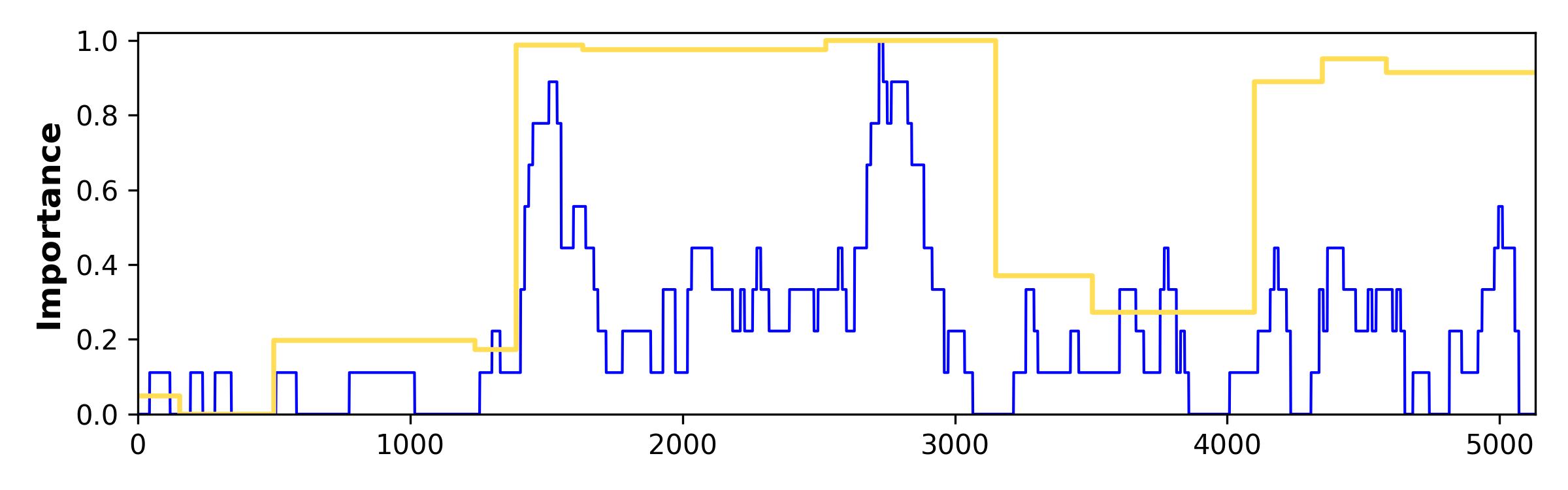} &
    \rowimg{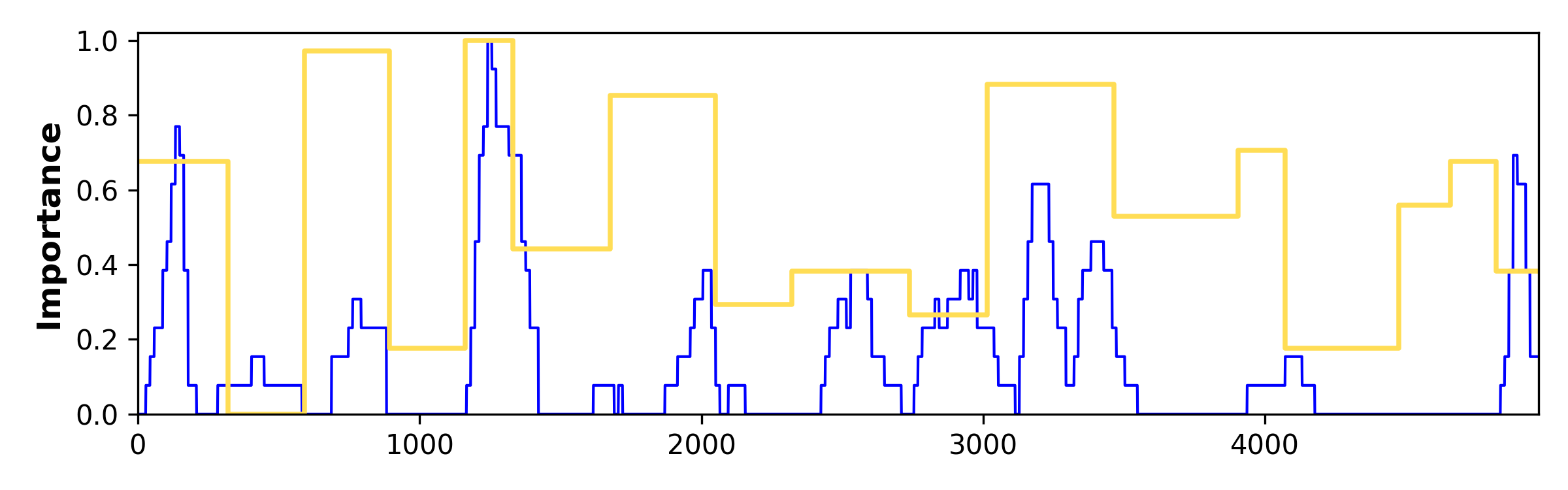} &
    \rowimg{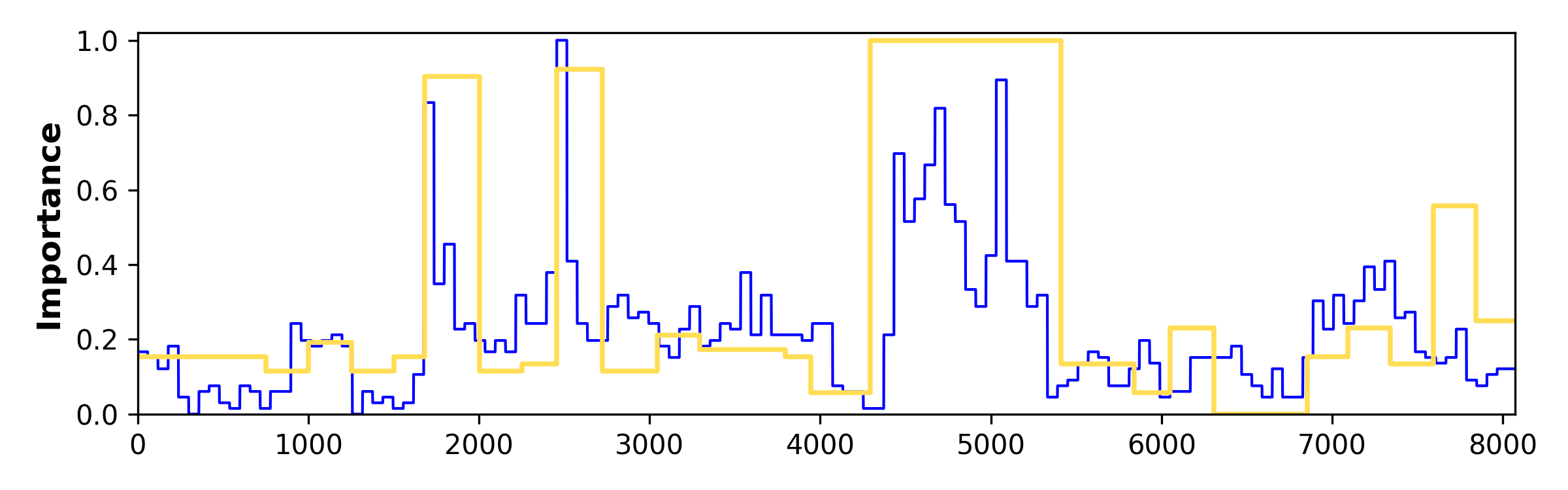} &
    \rowimg{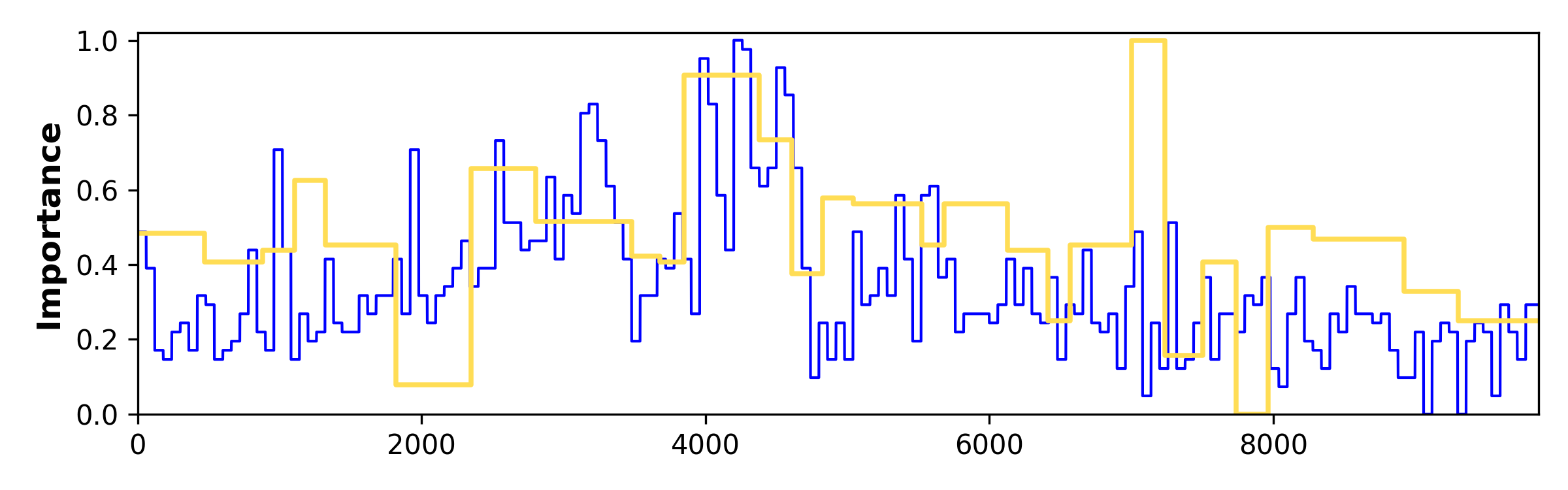} \\[-2pt]

    \makecell[c]{\textbf{(d) Normalized}\\[-2pt]
      \textbf{\& Smoothed}\\
      \textcolor{blue}{\footnotesize User annotations (mean)}\\[-2pt]
      \textcolor{myyellow}{\footnotesize Frame scores}}
    &
    \rowimg{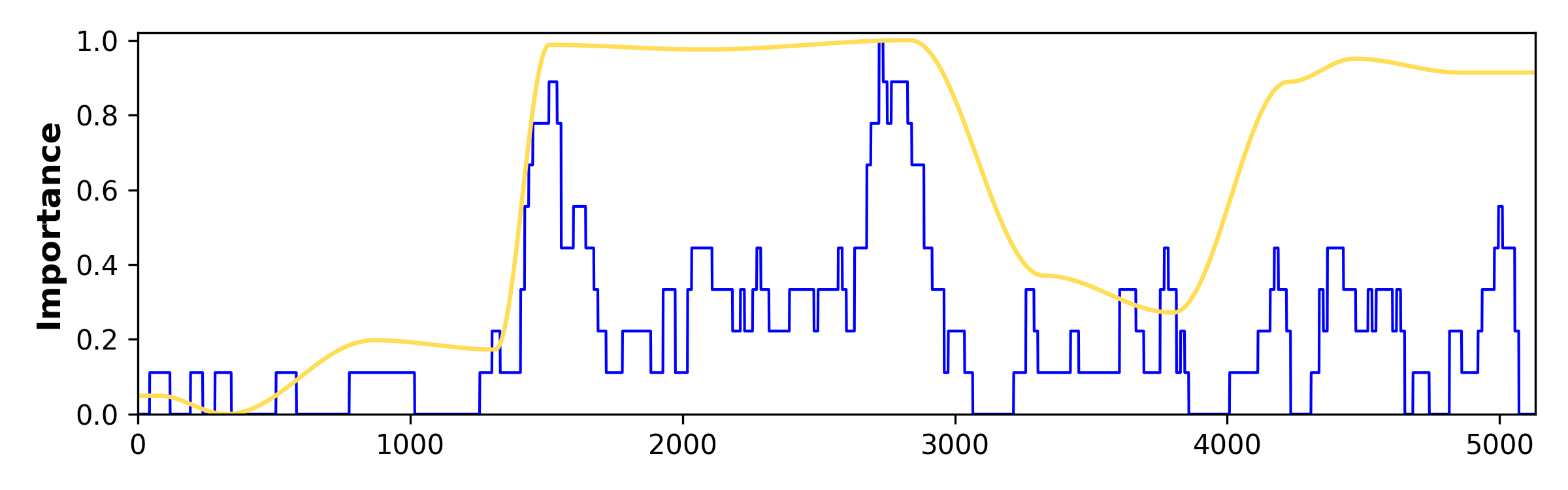} &
    \rowimg{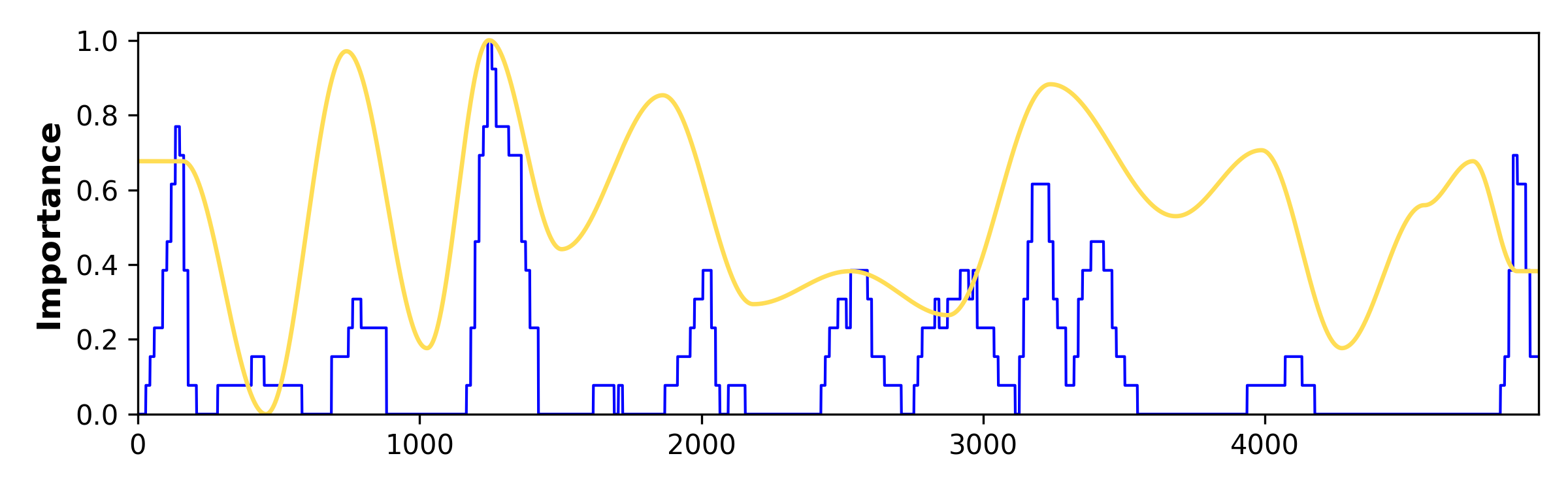} &
    \rowimg{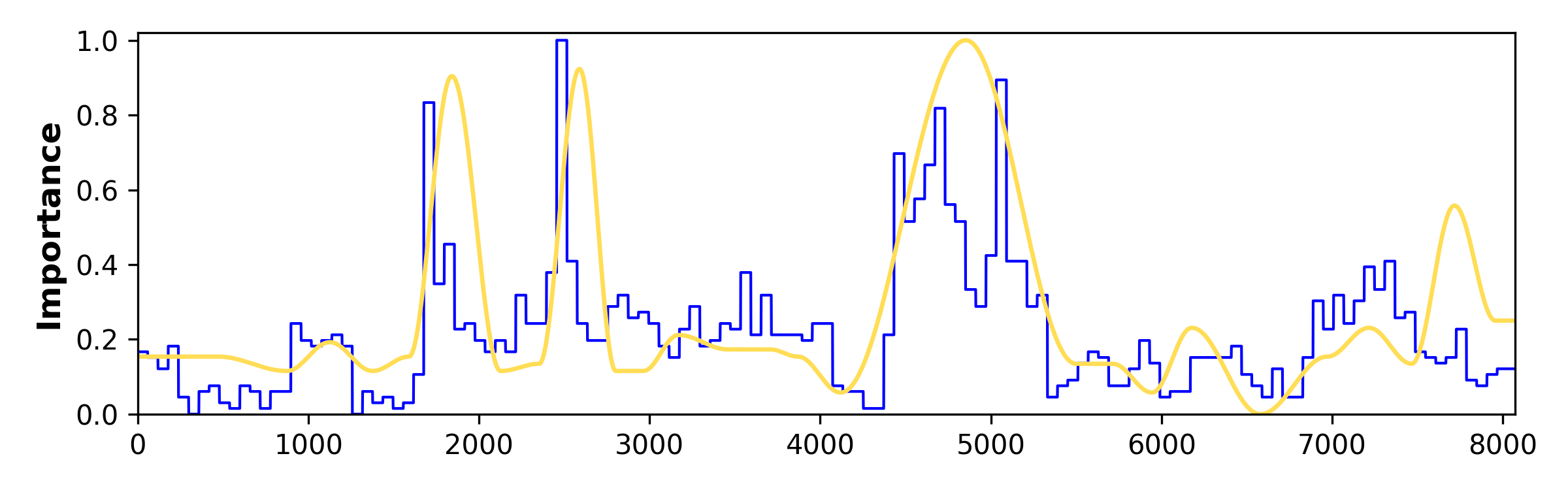} &
    \rowimg{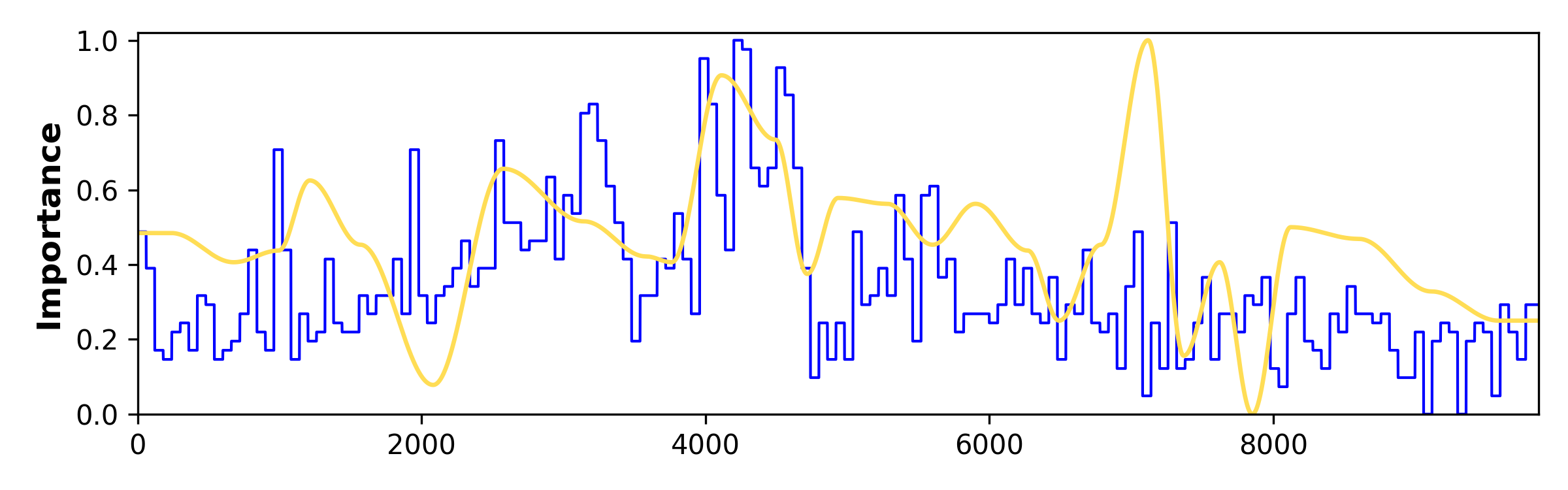} \\[-2pt]

    \makecell[c]{\textbf{(e) Frame-level}\\[-2pt]
      \textbf{scores}\\
      \textcolor{blue}{\footnotesize User annotations (mean)}\\[-2pt]
      \textcolor{myyellow}{\footnotesize Frame scores}}
    &
    \rowimg{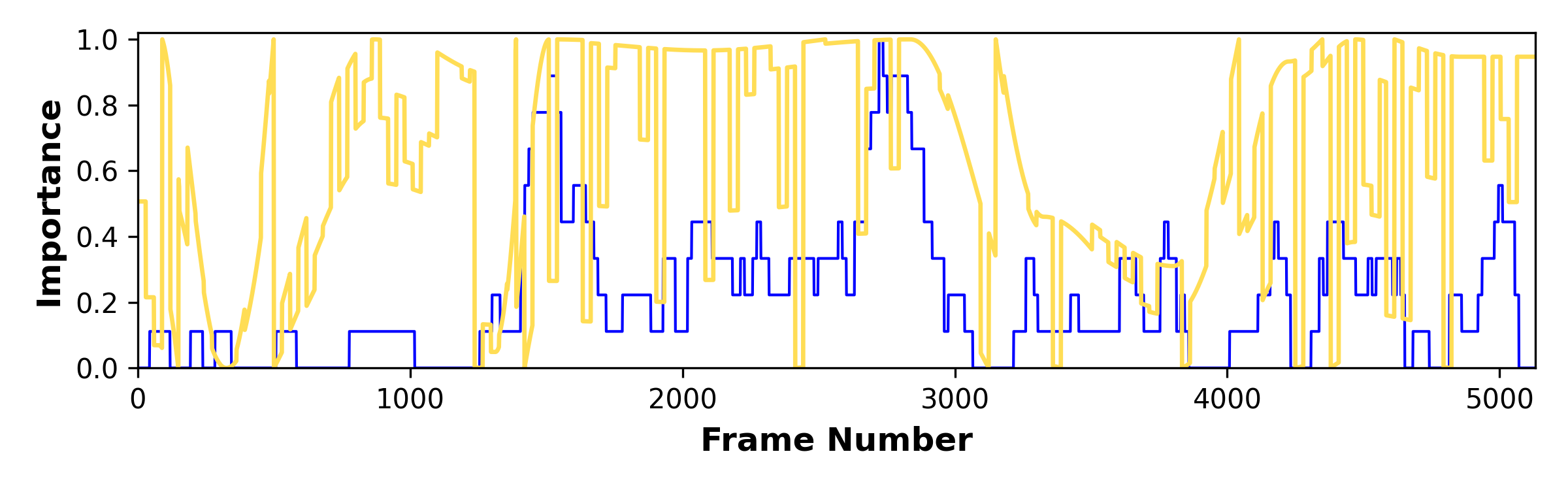} &
    \rowimg{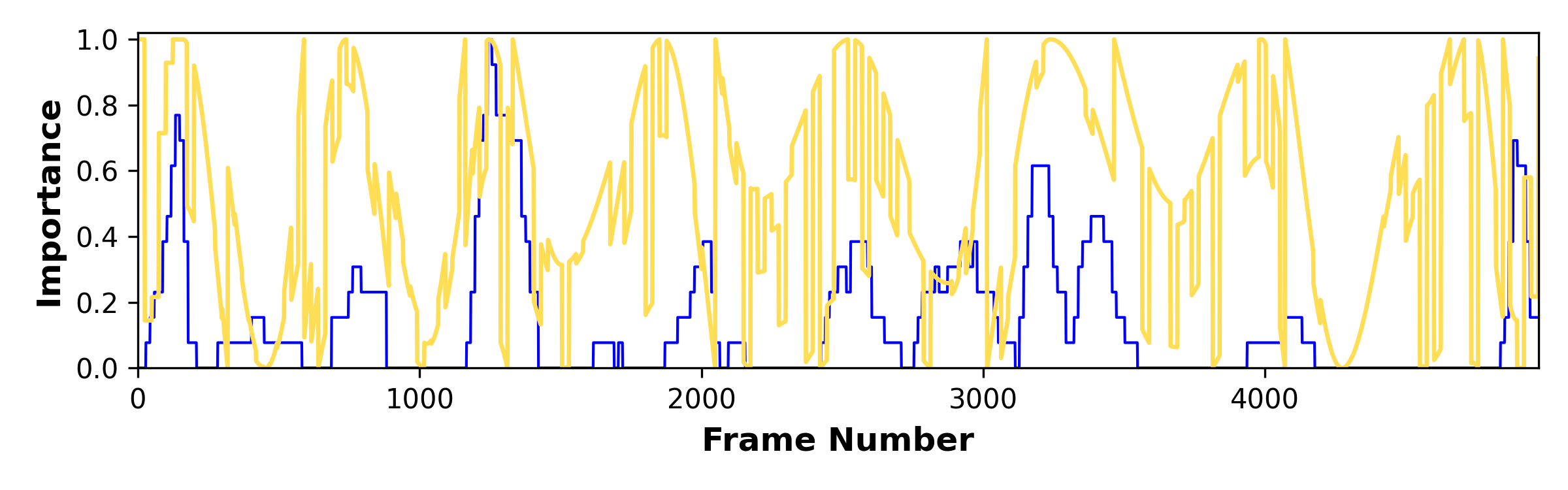} &
    \rowimg{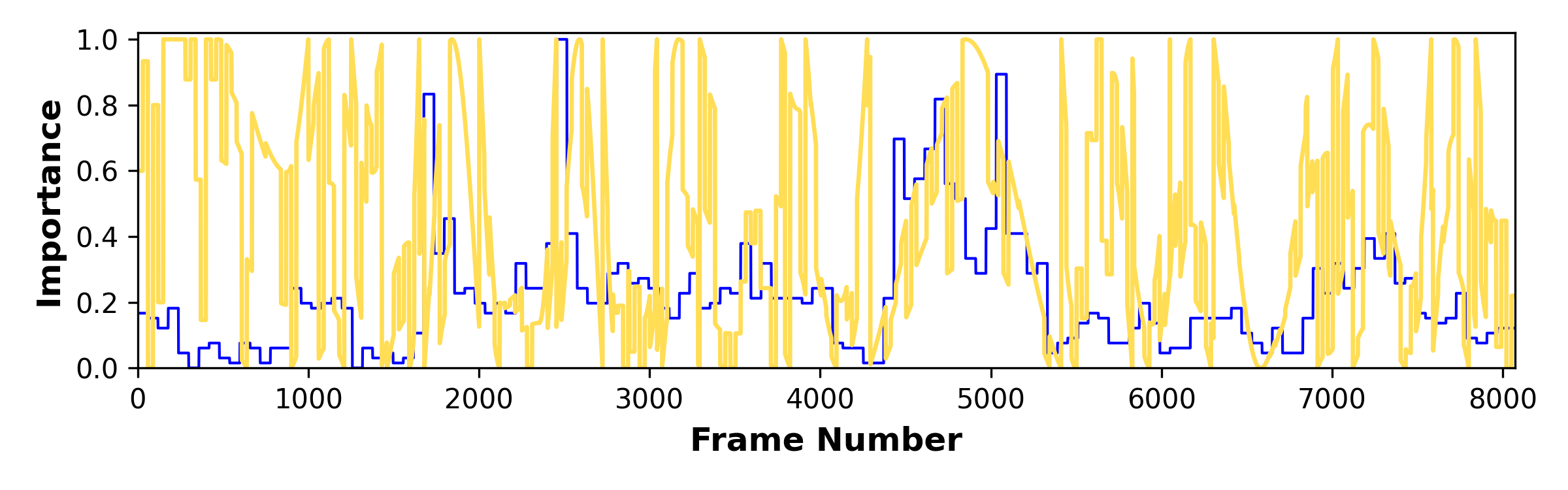} &
    \rowimg{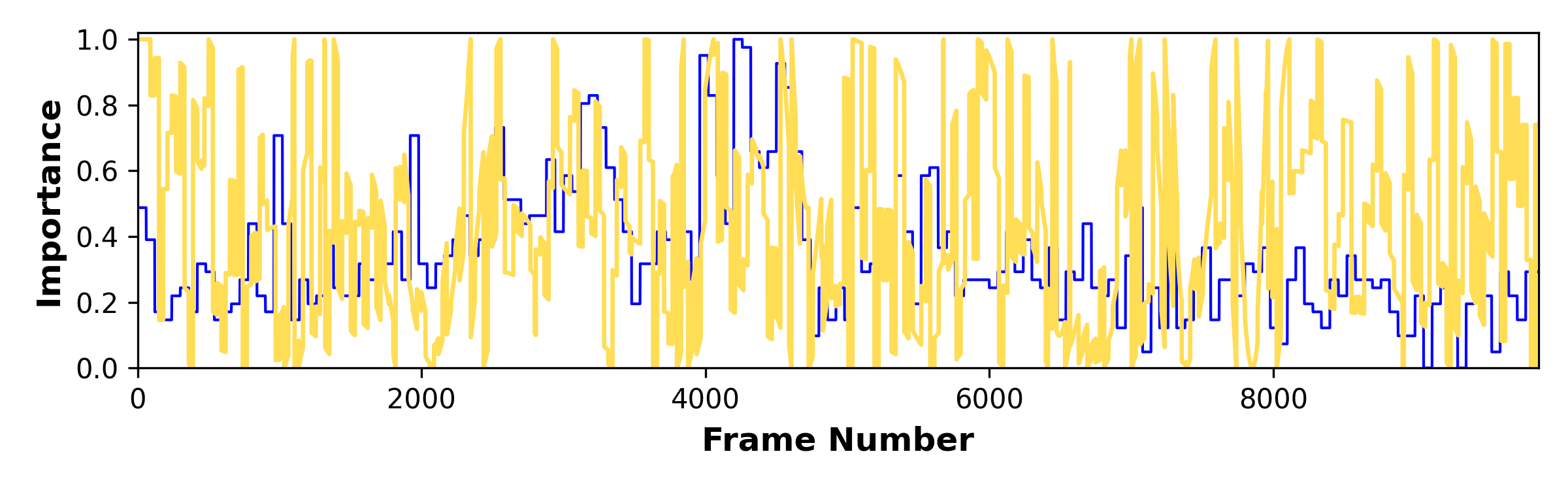} \\

  \end{tabular}

  \caption{the case study presents the complete step-by-step summarization pipeline, including: (a) \emph{Initial Scene Boundaries} detected from raw frame sequences, 
(b) \emph{Refined Scene Boundaries} adjusted based on embedding similarity, 
(c) \emph{Scene Scores} estimated by the language model using textual descriptions and user queries, 
(d) \emph{Normalized \& Smoothed Scores} obtained through temporal filtering, and 
(e) \emph{Frame-level Scores} representing the final output aligned with averaged user annotations. 
Blue lines denote user annotation means, and yellow curves indicate model-predicted frame importance scores. 
This visualization highlights how the proposed framework progressively refines temporal segmentation and scoring to achieve coherent, human-aligned video summaries.}
  \label{fig:pipeline_example}
\end{figure*}

\subsection{Case Study}
To qualitatively examine how our model generates summaries across different stages of processing, 
we conduct a case study on representative samples from both datasets. 
Specifically, two videos are selected from the \textbf{SumMe} dataset 
(\textit{Bus-in-Rock-Tunnel} and \textit{Eiffel-Tower}) 
and two from the \textbf{TVSum} dataset 
(\textit{qqR6AEXwxoQ} and \textit{vdmoEJ5YbrQ}). 
These samples cover diverse content types and event structures, making them suitable for observing temporal consistency and human–model alignment.
As illustrated in Figure~\ref{fig:pipeline_example}

\textbf{Results on SumMe.} As shown in Figure~\ref{fig:pipeline_example}, the two SumMe videos (\textit{Bus-in-Rock-Tunnel} and \textit{Eiffel-Tower}) 
demonstrate how our model progressively refines segmentation and scoring to match human annotations. 
In \textit{Bus-in-Rock-Tunnel}, the refined scene boundaries (b) remove redundant short segments from the initial detection (a), producing more stable divisions aligned with visual and semantic changes. The scene-level scores (c) and the smoothed frame-level predictions (d–e) closely follow the annotated importance peaks, particularly around key motion events such as the vehicle entering or exiting the tunnel. 
Minor deviations appear near transitions, where smoothing slightly broadens narrow human-labeled peaks, 
leading to higher recall but a small trade-off in precision—consistent with the quantitative results in Table~\ref{tab:ablation_tvsum_summe}. 
For \textit{Eiffel-Tower}, the model assigns higher scores to segments featuring clear interactions or architectural detail while down-weighting repetitive panoramic shots. 
This pattern shows that our rubric emphasizes concrete actions and visible detail over static scenery, resulting in more consistent precision gains on SumMe.

\textbf{Results on TVSum.} 
In contrast, the two TVSum examples (\textit{qqR6AEXwxoQ} and \textit{vdmoEJ5YbrQ}) display smoother annotation profiles and more stable segment transitions. 
Because TVSum videos have higher consistency in human labeling, the refined boundaries in (b) show only minor adjustments. 
The model’s predicted scores (yellow) align almost one-to-one with user annotations (blue), 
capturing most salient peaks while maintaining temporal coherence across the video. 
In \textit{qqR6AEXwxoQ}, both peak positions and amplitudes are well matched, confirming high precision and recall alignment. 
For \textit{vdmoEJ5YbrQ}, which contains frequent transitions and short motion bursts, the model effectively suppresses noise through smoothing 
while retaining the dominant peaks corresponding to meaningful actions.\\
Overall, the predictions on TVSum exhibit higher temporal stability and stronger correspondence with human annotations 
than those on SumMe, explaining the higher F1-score (63.05 vs.\ 57.58) reported in our quantitative evaluation.

\begin{figure*}[t]
  \centering
  \setlength{\tabcolsep}{0pt}
  \renewcommand{\arraystretch}{1.0}

  \begin{subfigure}{\linewidth}
    \centering
    \includegraphics[width=0.95\linewidth]{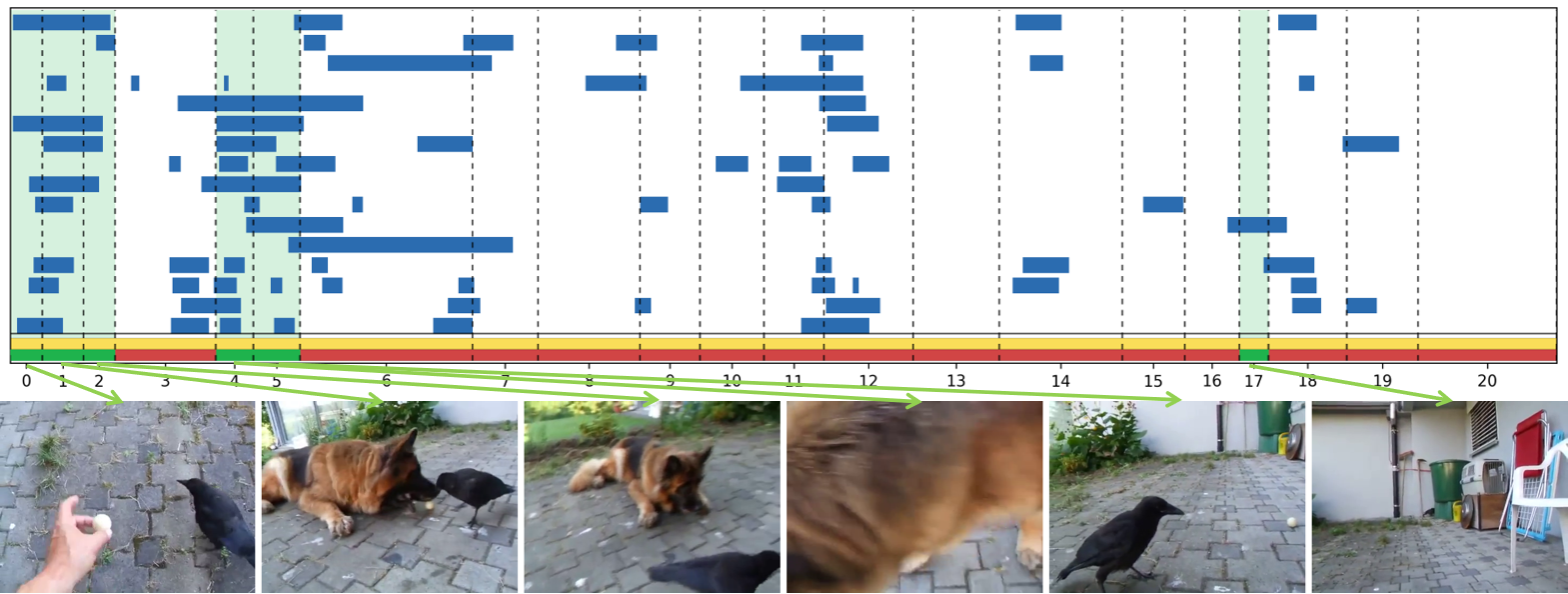}
    \caption{Visualization of the video summary for the video titled ``playing\_ball'' utilizing our model from \textbf{SumMe} dataset.}
    \label{fig:vis_summe}
  \end{subfigure}

  \vspace{0.5em}

  \begin{subfigure}{\linewidth}
    \centering
    \includegraphics[width=0.95\linewidth]{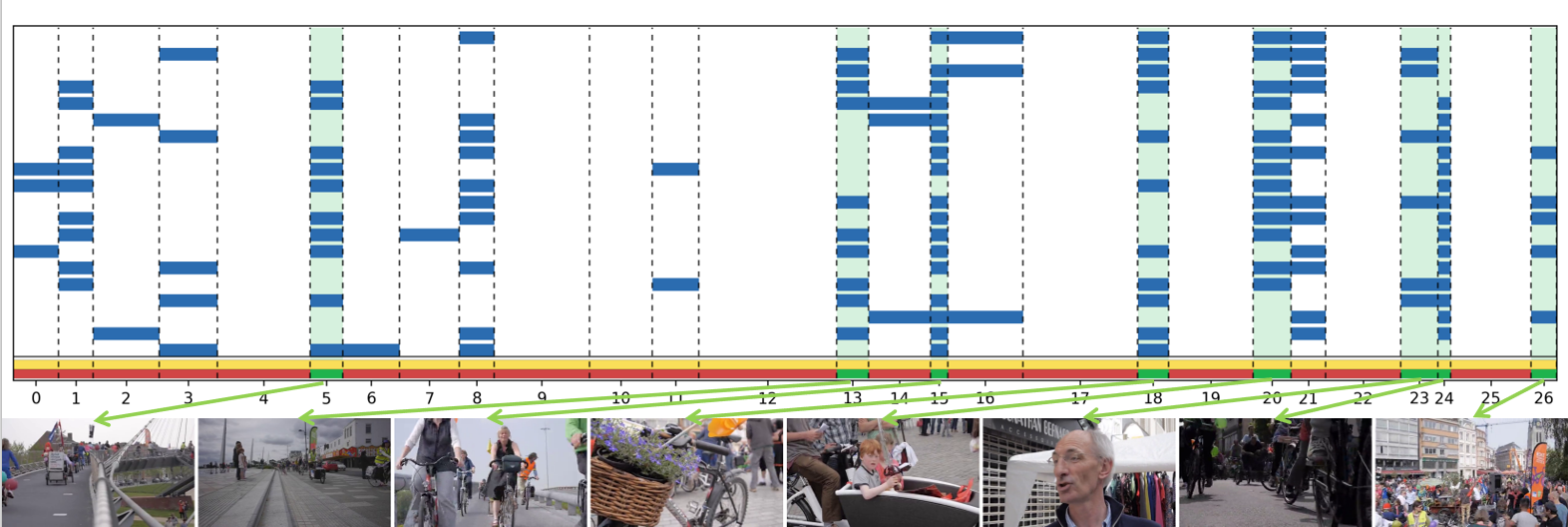}
    \caption{Visualization of the video summary for the video titled ``4wU\_LUjG5Ic'' utilizing our model from \protect\textbf{TVSum} dataset.}
    \label{fig:vis_tvsum}
  \end{subfigure}

  \caption{
    \textbf{Visualization of the selections made by our model, user summaries, and KTS segments.}~\cite{potapov2014category}
    The blue bars represent human frame selections for each video, with each row indicating a distinct user summary.
    The yellow bars denote ground-truth fragments included in the final summary.
    The green bars show the frames selected by our model, and the red bars mark unselected frames.
    Representative video frames from each selected shot are included for intuitive understanding.
  }
  \label{fig:qualitative_vis}
\end{figure*}
\subsection{Qualitative Visualization}

To better understand how the model aligns with human annotations and handles temporal segmentation, we conduct a detailed qualitative visualization on representative samples from both datasets, as shown in Figure~\ref{fig:qualitative_vis}. 
In each plot, blue bars represent individual user selections (one per row), green bars denote the frames selected by our model, and red regions indicate frames that were not selected. 
Dashed vertical lines mark the KTS scene boundaries, which divide the video into structural segments.

\textbf{(a) SumMe: playing\_ball.}
The SumMe sample exhibits a highly diverse annotation pattern, with blue bars forming multiple dense clusters separated by scattered, short individual selections. 
Within these clusters, users often disagree on the exact start and end points, reflecting high subjectivity in defining important moments. 
Our model’s selections (green) are concentrated in the middle of these high-density clusters, capturing the consensus portions where most users overlap. 
In low-consensus regions—where blue annotations appear fragmented or isolated—the model largely refrains from selection, leaving extended red zones. 
This behavior demonstrates the model’s ability to focus on representative, widely agreed-upon content while suppressing redundant or inconsistent user preferences. 
At the boundaries of dense clusters, thin red gaps appear, indicating that the model intentionally trims off inconsistent fringe frames to maintain compactness and temporal coherence. 
Overall, this visualization shows that in the presence of strong inter-user variability, our model prioritizes consensus regions and avoids scattered outliers, consistent with the lower annotation consistency observed in the SumMe dataset.

\textbf{(b) TVSum: 4wU\_LUjG5Ic.}
In contrast, the TVSum example presents smoother, more coherent annotation patterns. 
Here, blue bars form wide, stable vertical bands across multiple users, suggesting strong agreement on salient segments. 
The model’s green selections align almost perfectly with these dense blue bands, maintaining continuous coverage of core moments while clearly separating between distinct events. 
The red regions primarily occupy the intervals between clusters, showing that the model effectively suppresses transitional or low-activity intervals. 
A few small red gaps also appear at the edges of green segments, reflecting a deliberate tendency to keep summaries concise and avoid overlap between adjacent events. 
Visually, the predicted selections appear more stable and regular than in SumMe, reflecting higher temporal consistency and stronger alignment with collective user preferences—consistent with the superior F1 performance on TVSum (63.05).

\textbf{Cross-dataset comparison.}
Across both datasets, distinct behavioral patterns emerge. 
In TVSum, where user annotations are more consistent, the model tightly aligns with blue consensus bands, producing clear and continuous summaries. 
In SumMe, where annotations are more fragmented, the model exhibits stronger redundancy control—focusing on the central consensus regions while suppressing dispersed, low-agreement intervals. 
In both cases, the green selections effectively concentrate within areas of high human consensus, while red regions maintain separation across scene boundaries. 
These observations confirm that our approach achieves human-aligned, compact, and temporally coherent summaries, adapting its selection behavior to the annotation consistency of each dataset.
\vspace{-0.2cm}
\section{Conclusion}
This paper presents a pseudo-labeled, rubric-guided, and prompt-driven zero-shot video summarization framework that bridges large language models (LLMs) with structured semantic reasoning. 
Without any supervised training, the proposed method achieves competitive and stable performance across three benchmarks: \emph{SumMe}, \emph{TVSum}, and the query-focused long-video dataset \emph{QFVS}. 
Specifically, our approach attains F1 scores of \textbf{57.58}, \textbf{63.05}, and \textbf{53.79} on SumMe, TVSum, and QFVS, respectively, surpassing the zero-shot baseline by \textbf{+0.85}, \textbf{+0.84}, and \textbf{+0.37}. 
These results demonstrate the scalability of our rubric-guided prompting to both short-form and long-form, query-driven summarization scenarios. 
Ablation analyses confirm that pseudo-label reconstruction improves scoring consistency, while context-aware prompt design further enhances temporal coherence and redundancy suppression. 
Qualitative visualizations show that the generated summaries align closely with human consensus, capturing salient and narrative-rich moments. 
Overall, this work illustrates that language-driven reasoning provides a powerful, interpretable, and training-free alternative to traditional data-dependent summarization models.
\vspace{-0.1cm}
\section{Limitation}
Despite its effectiveness, our framework has several limitations. 
First, the reliability of the pseudo-labeled rubric depends heavily on the quality and diversity of the initial annotations; biased or low-consensus labels may propagate dataset-specific tendencies and affect generalization. 
Second, although the framework eliminates supervised training, inference with large language models remains computationally demanding, limiting scalability for extremely long or real-time videos. 
Finally, the current design focuses solely on visual information and does not yet incorporate multimodal cues such as audio, subtitles, or textual metadata, which could further enhance semantic understanding, especially for query-focused summarization. 
Future work will explore confidence-aware pseudo labeling, lightweight model distillation, and multimodal integration \cite{wen2024learning, diao-etal-2024-learning} to improve efficiency, robustness, and cross-dataset generalization.

\section*{Acknowledgment}
We acknowledge the use of GPT-5~\cite{openai2025gpt5} for grammar correction, typo fixing, and partial sentence rephrasing to improve the clarity and readability of this paper.

\bibliographystyle{ACM-Reference-Format}
\bibliography{Reference}
\end{document}